\documentclass{naturep}

\usepackage{amssymb}
\usepackage{amsmath}
\usepackage{graphicx}
\usepackage{algorithmicx}
\usepackage{algorithm}

\usepackage[noend]{algpseudocode}
\usepackage{bbm}
\usepackage{lineno}
\usepackage[margin=0.6in]{geometry}
\usepackage{ragged2e}
\usepackage{setspace}
\usepackage{longtable}
\usepackage{hyperref}
\usepackage{anyfontsize}
\usepackage{setspace}
\usepackage{float}
\usepackage{booktabs}
\usepackage{multirow}
\usepackage{xcolor}
\usepackage{blindtext}
\usepackage{tabularx}
\usepackage{caption}
\usepackage{subcaption}

\makeatletter
\let\saved@includegraphics\includegraphics
\AtBeginDocument{\let\includegraphics\saved@includegraphics}

\makeatother

\usepackage{enumitem, kantlipsum}

\title{\begin{flushleft}{\begin{spacing}{1} A Foundational Multimodal Vision Language AI Assistant for Human Pathology\end{spacing}}\end{flushleft}}

\begin{document}

\maketitle
\begin{spacing}{1.8}
\vspace{-15mm}
\noindent Ming Y. Lu$^{1,2,3,4,6\boldsymbol{\ddag}}$, Bowen Chen$^{1,2\boldsymbol{\ddag}}$, Drew F. K. Williamson$^{1,2,3\boldsymbol{\ddag}}$, Richard J. Chen$^{1,2,3,4,5}$, Kenji Ikamura$^{1,2}$, Georg Gerber$^{1}$, Ivy Liang$^{1,7}$, Long Phi Le$^{2}$, Tong Ding$^{1}$, Anil V Parwani$^{8}$, Faisal Mahmood$^{*1,2,3,4,9}$
\end{spacing}
\vspace{-6mm}
\begin{spacing}{1.4}
\begin{affiliations}
 \item Department of Pathology, Brigham and Women's Hospital, Harvard Medical School, Boston, MA
 \item Department of Pathology, Massachusetts General Hospital, Harvard Medical School, Boston, MA
 \item Cancer Program, Broad Institute of Harvard and MIT, Cambridge, MA 
 \item Cancer Data Science Program, Dana-Farber Cancer Institute, Boston, MA
 \item Department of Biomedical Informatics, Harvard Medical School, Boston, MA
 \item Electrical Engineering and Computer Science, Massachusetts Institute of Technology (MIT), Cambridge, MA
 \item Harvard John A. Paulson School of Engineering And Applied Sciences, Harvard University, Cambridge, MA
 \item Department of Pathology, Wexner Medical Center, Ohio State University, Columbus, OH
 \item Harvard Data Science Initiative, Harvard University, Cambridge, MA
 \\$\boldsymbol{\ddag}$ Contributed Equally
 \\\textbf{*Corresponding author}: Faisal Mahmood (faisalmahmood@bwh.harvard.edu)
\end{affiliations}
\end{spacing}

\vspace{-5mm}
\begin{spacing}{1.2}
\noindent \textbf{Abstract}\\
The field of computational pathology has witnessed remarkable progress in the development of both task-specific predictive models and task-agnostic self-supervised vision encoders. However, despite the explosive growth of generative artificial intelligence (AI), there has been limited study on building general purpose, multimodal AI assistants tailored to pathology. Here we present PathChat, a vision-language generalist AI assistant for human pathology using an in-house developed foundational vision encoder pretrained on 100 million histology images from over 100,000 patient cases and 1.18 million pathology image-caption pairs. The vision encoder is then combined with a pretrained large language model and the whole system is finetuned on over 250,000 diverse disease agnostic visual language instructions. We compare PathChat against several multimodal vision language AI assistants as well as GPT4V, which powers the commercially available multimodal general purpose AI assistant ChatGPT-4. When relevant clinical context is provided with the histology image, PathChat achieved a diagnostic accuracy of 87\% on multiple-choice questions based on publicly available  cases of diverse tissue origins and disease models. Additionally, using open-ended questions and human expert evaluation, we found that overall PathChat produced more accurate and pathologist-preferable responses to diverse queries related to pathology. As an interactive and general vision language AI assistant that can flexibly handle both visual and natural language inputs, PathChat can potentially find impactful applications in pathology education, research, and human-in-the-loop clinical decision making. \\
\textbf{Video demo}: \href{https://www.dropbox.com/scl/fi/ipy9tbqqu9sub445wb749/pathchat_demo_v1.mp4?rlkey=e61alehlz62m7eimvyzctwvh2&dl=0}{Dropbox link}, \href{https://drive.google.com/file/d/1-k0fpGTKnjagJggvRDeJTUYMJEXwlmFE/view?usp=sharing}{Google Drive link}

\end{spacing}

\newpage
% \linenumbers

\begin{spacing}{1.35}
\noindent\textbf{\large{Introduction}} 

The field of computational pathology has witnessed a remarkable transformation in recent years, propelled by the convergence of several key trends including increased availability and institutional adoption of digital slide scanning, rapid progress in artificial intelligence (AI) research, increasing accessibility of large datasets, and substantial high-performance computing resources\cite{madabhushi2016image,shmatko2022artificial,song2023artificial,bera2019artificial,heinz2022future, cui2021artificial, abels2019computational, waqas2023revolutionizing, lipkova2022artificial}. With varying degrees of success, researchers have leveraged deep learning to address a diverse range of tasks, including cancer subtyping\cite{coudray2018classification, lu2021data} and grading\cite{bulten2020automated, nagpal2019development}, metastasis detection\cite{huang2022deep,campanella2019clinical, bejnordi2017diagnostic}, survival \cite{beck2011systematic,mobadersany2018predicting, porpoise, lee2022derivation, courtiol2019deep, lu2020prognostic, amgad2023population, boehm2022multimodal} and response-to-treatment prediction\cite{sammut2022multi, vanguri2022multimodal,huang2023artificial}, tumor site of origin prediction\cite{lu2021ai, zhu2023accurate}, image search\cite{sish, yottixel, smily, wang2023retccl}, mutation prediction and biomarker screening\cite{kather2020pan, fu2020pan, saldanha2023self,wagner2023transformer}, and more\cite{yala2022optimizing,zhou2023multi,laleh2022benchmarking,graham2019hover,graham2023one}. At the same time, general purpose vision encoder models\cite{dino, dinov2, mae, simclr, moco, ibot}, which are trained on vast datasets of unlabeled histopathology images and can serve as versatile task-agnostic model backbones\cite{uni, ctranspath, google_ssl, virchow, remedis, fuchs_ssl, lunit_ssl, ciga_ssl}, are paving the way for further improvements across many tasks in computational pathology, both in performance and label efficiency. 

However, the aforementioned developments in computational pathology do not yet reflect the important roles of natural language in pathology, as a key to unlocking rich, diverse sources of accumulated human medical knowledge, as a potential signal for model supervision, and as a unified medium for facilitating intuitive interaction between powerful AI models and end users. Notably, in general machine learning, representative works demonstrated that large-scale vision language representation learning can augment vision-only AI models with new capabilities including zero-shot image recognition\cite{clip, align} and text-to-image retrieval\cite{unicoder_vl, vilbert, uniter}. Depending on the architectural design and training data and objectives, visual language pretrained systems\cite{oscar, git, vinvl, lemon, beitv3, coca, blip, albef} can also often be finetuned for tailored tasks ranging from visual question answering and image captioning to object detection and semantic segmentation. In medical imaging and computational pathology, researchers have recently begun to harness diverse sources\cite{schaumberg2020interpretable, laion, convirt, roco_dataset, plip, biomedclip, arch, quilt} of paired biomedical images and captions or reports for visual language pretraining\cite{pmc-clip, biomedclip}, including the development of CLIP\cite{clip}-like models tailored for specific domains such as pathology\cite{plip, conch, mizero, quilt} and radiology\cite{tiu2022expert, convirt, gloria, biovl, pubmedclip}. In computational pathology, a few works have shown promising zero-shot performance in select diagnostic and retrieval tasks\cite{plip, conch, quilt}, while other works also experimented with designing specialized models for biomedical visual question answering or captioning\cite{zhang2023pathnarratives,tsuneki2022inference,zhang2020evaluating,naseem2022vision,he2021towards}. However, these models are not yet ready to serve as interactive assistants for pathologists, researchers using pathology image data, or pathology trainees. This is because those models such as CLIP which have been trained using pathology data are not also trained to understand and follow diverse and often complex instructions of users and have limited to no ability to generate coherent responses in natural language grounded in factual knowledge. 

Following the rise of large language models\cite{vaswani2017attention, gpt1, gpt2, gpt3, chinchilla, wei2022emergent, instructgpt, llama, llama2, t5, zhao2023survey, palm, palm2, medflamingo} (LLMs), rapid advances in multimodal large language models\cite{li2023multimodal,llava, flamingo} (MLLMs) and the broader field of generative AI\cite{moor2023foundation} are poised to open a new frontier for computational pathology, one which emphasizes natural language and human interaction as key components of AI model design and user experience, in addition to powerful visual processing capabilities. General purpose multimodal generative AI products, notably ChatGPT-4, are maturing, and have demonstrated impressive capabilities on a wide range of routine, creative, as well as professional use cases\cite{gpt4_exp}, including coding, writing, summarization, data analysis, question answering, translation, and even image generation, all while being accessible through an intuitive and interactive user interface. While there have been attempts to investigate their performance on answering medicine-related queries\cite{clinical_bert, med-palm2, med-palm-m, nyutron, ayers2023comparing, gpt4_medprompt, gpt4_medchallenge, buckley2023accuracy}, their capability to assist professionals and researchers in the highly specialized but important sub-field of anatomic pathology remains relatively unexplored\cite{pathasst, llava-med, med-palm-m, wu2023can, oon2023bridging}. However, the potential applications of an interactive, multimodal AI assistant for pathology are immense. The ability to understand and respond to complex queries in natural language in theory could enable such an assistant to serve as a helpful companion across various stages of human-in-the-loop clinical decision making, education, and research. For instance, in the clinic, the AI assistant might be able to ingest a histopathology image, provide an initial assessment of the morphological appearance, and identify potential features of malignancy. Subsequently, the pathologist or trainee could provide additional context about the underlying case, such as clinical parameters of the patient and the tissue site, and ask the model to suggest a differential diagnosis. If deemed reasonable, the user could then request helpful suggestions for ancillary testing and immunohistochemical stains to narrow down the differential. Finally, the results of such tests could also be provided to the model to make a final deduction to arrive at the diagnosis. In research, a multimodal AI model that can summarize the morphological features of large cohorts of histopathology images would potentially enable automated quantification and interpretation of morphological markers in large data cohorts. In medical education, an accurate, on-demand interactive AI companion could help democratize access to expert-level guidance and training in pathology, thereby narrowing the gap between regional disparities in healthcare provision.

\noindent\textbf{Vision language AI assistant for human pathology}\\
In this article, we develop a vision language interactive AI assistant for human pathology powered by a custom, finetuned multimodal large language model (MLLM). To build an MLLM-based vision language AI assistant that can reason over both visual and natural language inputs, we begin with UNI\cite{uni}, a state-of-the-art (SOTA) vision-only foundational encoder model pretrained on over 100 million histology images from over 100 thousand slides using self-supervised learning. We perform further vision language pretraining on the UNI encoder with 1.18 million pathology image caption pairs to align its image representation space with that of pathology text\cite{conch}. The resulting vision encoder, CONCH-Large, is subsequently
connected to a 13 billion parameter, pretrained LLM\cite{llama2} via the multimodal projector module to form
the complete MLLM architecture. The MLLM is finally finetuned via a curated dataset of over 250 thousand instructions to build PathChat (\textbf{Figure 1}), a visual language AI assistant that can understand pathology images and text and respond to complex pathology-related queries. More details about data curation and model training can be found in PathChat dataset curation and PathChat model design and training section of Methods respectively.

We dmonstrate the capabilities of PathChat in various applications including analysis of pathology cases from diverse organ sites and practices (\textbf{Figures 2 and 3}). Additionally, we contribute, to the best of our knowledge, the first high quality open-ended pathology visual question answering benchmark curated with expert supervision (see the \textbf{PathQABench: an expert-curated pathology question-answering benchmark} section of \textbf{Methods} for more details), making it suitable for evaluating the performance of multimodal large language models in pathology and fulfilling a critical need amidst rapid advances of visual language multimodal AI research. We evaluate our custom built vision language AI assistant for pathology, named PathChat, against both LLaVA\cite{llava}, a state-of-the-art (SOTA) general domain open-source MLLM as well as LLaVA-Med\cite{llava-med}, which has been tailored to the biomedical domain. We also compare against a SOTA commercial solution, ChatGPT-4 (powered by GPT4V), despite our model being significantly smaller and cheaper to serve. More details about model design and training can be found in the \textbf{PathChat model design and training section} of \textbf{Methods} with hyperparamters described in \textbf{Extended Data Tables 1-3}.

% Bard\footnote{\href{https://bard.google.com/chat/}{bard.google.com/chat/}}

\begin{figure*}[t]
% \vspace{-9mm}
\centering
\includegraphics[width=\textwidth]{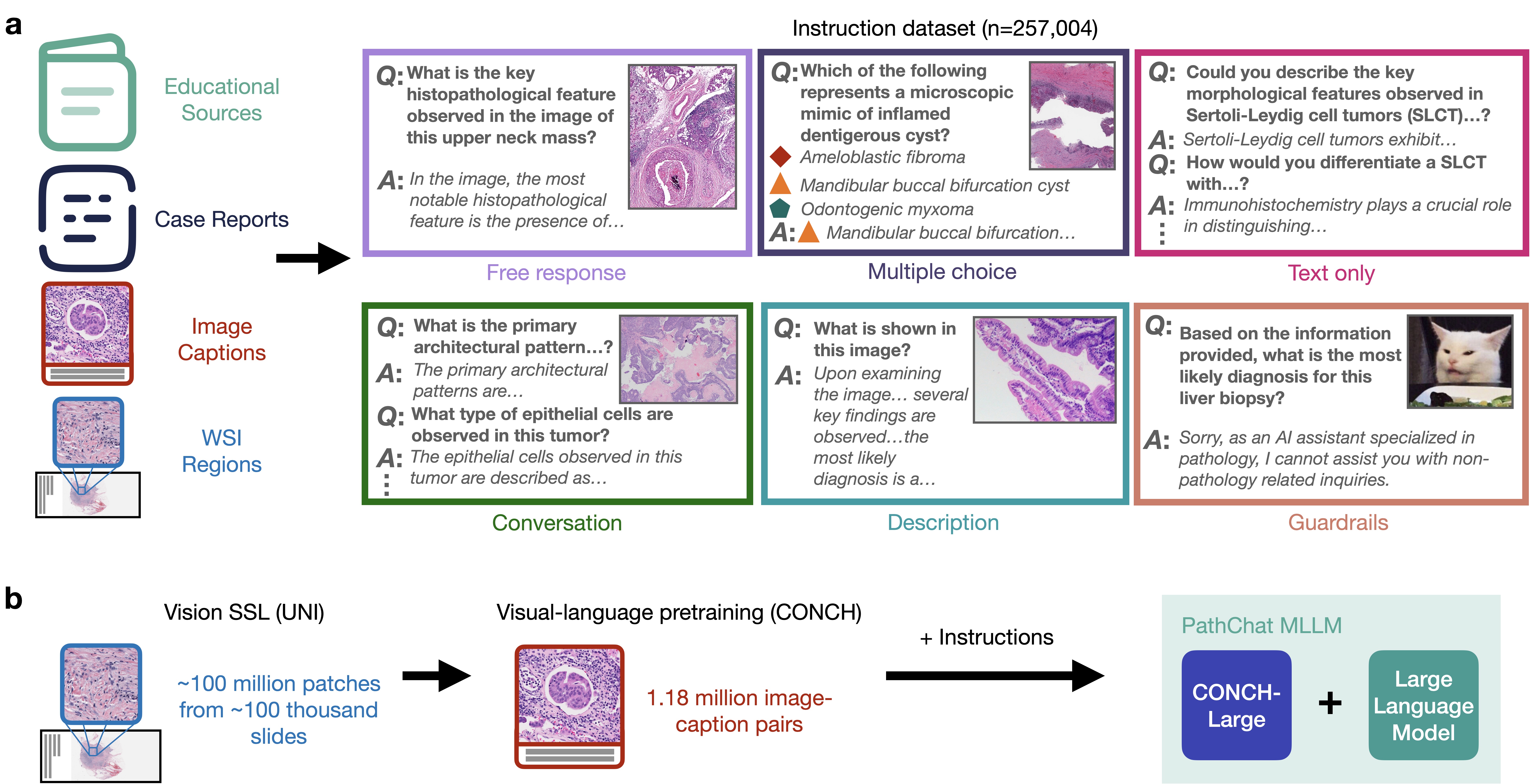}
\caption*{\textbf{Figure 1: Instruction-following dataset curation and PathChat overview.} \textbf{a.} We curated the currently largest instruction finetuning dataset specialized for the domain of pathology, consisting of 257k instructions and corresponding responses covering varied formats (\textit{e.g.} multi-turn conversations, multiple-choice questions, short answers; see \textbf{Extended Data Figure 1} for complete examples) from diverse sources.
\textbf{b.} To build an MLLM-based vision language AI assistant that can reason over visual and natural language inputs, we begin with a SOTA vision-only self-supervised pretrained foundation encoder model, UNI, and perform further vision language pretraining analogous to CONCH. The resulting vision encoder, CONCH-Large, is subsequently connected to a 13 billion parameter, pretrained LLM via a multimodal projector module (not shown) to form the complete MLLM architecture. The MLLM is finetuned via the curated instruction-following dataset to build PathChat, a visual language AI assistant specialized for human pathology. More details about data curation and model training can be found in \textbf{PathChat dataset curation} and \textbf{PathChat model design and training} section of \textbf{Methods} respectively.}
\end{figure*}

\noindent\textbf{\large{Results}}

\noindent\textbf{Performance on multiple-choice diagnostic questions} \\
We began by assessing the capability of our PathChat MLLM to directly make a diagnosis from histology images. For this purpose, a board-certified pathologist manually selected salient regions of interest (ROIs) from routine H\&E diagnostic whole slide images (WSIs) from both the TCGA and our in-house pathology archive (both of which are completely withheld from model pretraining or finetuning). In total, the questions cover 29 diagnoses from 9 different major pathology practices and organ sites (\textbf{Extended Data Tables 4-5}). For each organ system, the pathologist selected a set of 10 possible answers that encompasses the correct answers for all questions within that organ system as well as other relatively common diagnoses within that organ system (\textbf{Extended Data Table 6}). 
% \clearpage
For each question, we consider two evaluation strategies.
In the first, the image-only setting, the model is presented with only the image and the multiple choice question as input. In the second, the image with clinical context setting, designed to more closely mimic a real-world diagnostic workflow, additional relevant clinical context is provided together with the histology image, including information that may include patient age, sex, clinical history, and radiology findings as appropriate for the clinical case. In both settings, the model is assessed for its ability to accurately select the ground truth diagnosis from the set of possible options. We provide an illustrative example of the complete model input in \textbf{Figure 2a}. On all cases (denoted as ``Combined" in \textbf{Figure 2b}), we compare PathChat against LLaVA 1.5, a SOTA general purpose visual language chatbot assistant, as well as LLaVA-Med, a specialized version of LLaVA finetuned for answering biomedical related queries. Additionally, on the subset of 23 cases derived from publicly available WSIs (denoted as ``PathQABench-Public"), in addition to LLavA 1.5 and LLaVA-Med, we also compare against GPT4V, which powers ChatGPT4, the current best in class, vision capability enabled commercial AI assistant developed by OpenAI. All models were evaluated ``as is" without any additional task-specific finetuning, consistent with the paradigm of zero-shot transfer.

In both evaluation settings (image only and image with clinical context), PathChat convincingly outperforms the open-source baselines LLaVA 1.5 and LLaVA-Med in terms of diagnostic accuracy (\textbf{Figure 2b}, \textbf{Extended Data Tables 7-9}). In the image only evaluation setting, PathChat scored a 70.8\% accuracy (+50\% \textit{vs.} LLaVA 1.5 and +52.1\% \textit{vs.} LLaVA-Med) on the full combined benchmark. In line with expectation, the performance of PathChat further improves to 81.2\% accuracy (+54.1\% \textit{vs.} LLaVA 1.5 and LLaVA-Med) when additional useful clinical context is provided. Specifically, we note that the addition of clinical context consistently improves the accuracy of PathChat both on the private in-house cases (PathQABench-Private, +16\%) and the public TCGA cases (PathQABench-Public, +4.4\%). These findings suggest that PathChat can effectively and flexibly leverage multimodal information for more accurate diagnosis of histology images by simply providing such additional non-visual information in plain natural language without specialized data processing. 

Additionally, using PathQABench-Public, which only contains cases from the publicly available TCGA WSIs, we also compare our model against the GPT-4Vision (GPT4V) model. We observe that our domain-specific PathChat MLLM achieves higher diagnostic accuracy compared to GPT4V in both evaluation settings on the cases tested, although it is much more accurate than the open-source MLLMs tested, especially when clinical context is provided. Additionally, it is worth noting that guardrails appear to have been implemented into GPT4V to sometimes prevent it from addressing queries that require examination of medical images, and it will instead inform the user that it cannot provide a pathology interpretation and instead to consult a medical professional. In such cases, we make a maximum of 2 additional submissions for the same query for a total of up to 3 attempts (see \textbf{Evaluation of GPT4V} of the \textbf{Methods} section for more details). Following this evaluation protocol, we successfully queried GPT4V for all 23 PathQABench-Public images when clinical context is included, but only 12 out of 23 questions for the image only setting. An ultimately unsuccessful query was treated as incorrect since the response did not address the question. However, we also report performance on only the subset of questions that GPT4V successfully answered (\textbf{Extended Data Table 10}).

\begin{figure*}[h!]
% \vspace{-9mm}
\centering
\includegraphics[width=.9 \textwidth]{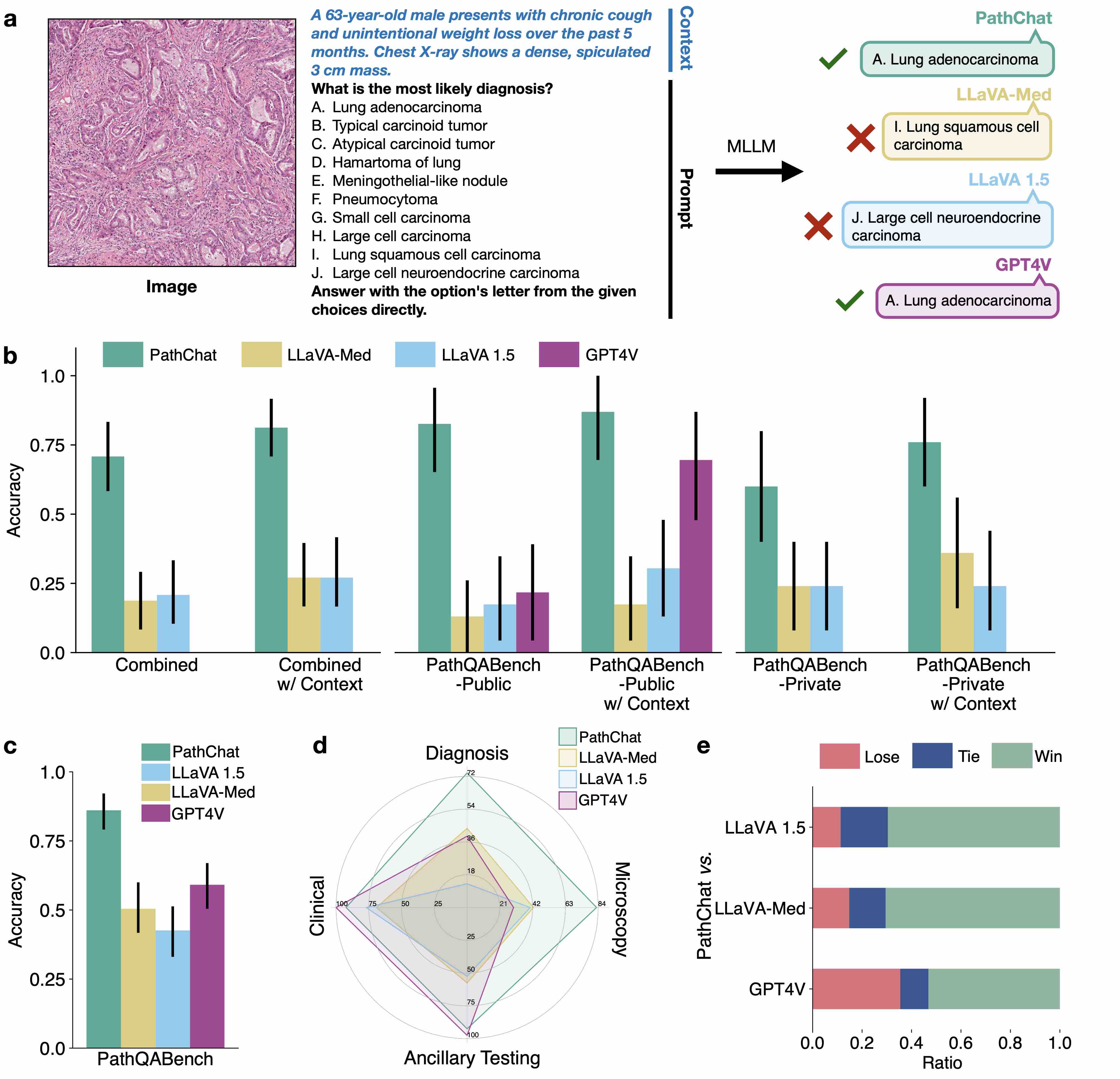}
\caption*{\textbf{Figure 2: Multiple choice evaluation of PathChat.} \textbf{a.} Illustrative example of a multiple-choice style diagnostic question. The input always includes a salient histology image ROI selected by a board-certified anatomic pathologist and the instruction to select the most likely diagnosis from a set of possible choices. In the image + clinical context evaluation setting that is designed to more closely mimic a real-world diagnostic workflow, additional relevant clinical context (designed by the pathologist, shown in blue) is provided together with the histology image and concatenated with the original instruction. \textbf{b.} Accuracy of MLLMs on multiple choice-style diagnostic questions. Note that we only compare against GPT4V on questions based on publicly available cases (PathQABench-Public). \textbf{c.} Accuracy of MLLMs on open-ended questions. \textbf{b, c.} Error bars represent 95\% confidence intervals. \textbf{d.} Accuracy on different categories of questions. \textbf{e.} Head-to-head records on open-ended questions for PathChat \textit{v.s.} other MLLMs. Lose: said model is ranked higher than PathChat; Tie: PathChat is tied with the model in ranking; Win: PathChat is ranked higher than the model.}  
\end{figure*}

\noindent\textbf{Performance on open-ended question answering} \\
Beyond multiple-choice diagnostic questions, it is valuable to assess the ability of PathChat and other MLLMs to generate coherent, reasonable, and clinically relevant responses to open-ended pathology-related inquiries (see \textbf{PathQABench: an expert-curated pathology question-answering benchmark} section of \textbf{Methods}). Based on cases from PathQABench-Public, a board-certified anatomic pathologist carefully open-ended questions targeting a broad spectrum of topics including microscopic image description, histologic grade and differentiation status, risk factors, prognosis, treatment, diagnosis, immunohistochemical (IHC) tests, molecular alterations, and other further testing. Similar to the multiple-choice evaluation, to mimic the real-world use case of a pathology AI assistant, each question is provided to models as is, without requiring any additional model or task-specific finetuning. The responses are then evaluated by an expert pathologist both in terms of accuracy (\textit{i.e.} a binary label of correct \textit{vs.} incorrect), as well as ranking (from best to worst, ties allowed) based on their relevance to the question, correctness, and whether it is supplemented with correct explanation or reasoning in a succinct manner (see \textbf{MLLM evaluation} section of \textbf{Methods} for more details and \textbf{Extended Data Figures 2-4} for illustrative examples of ranked model responses). Throughout the ranking process, the pathologist, who has not had prior interaction with each model, is also blinded to which model produced which response and the responses were additionally randomly shuffled for each question to further minimize potential bias towards specific models. 

Overall, we find that PathChat produces on average more accurate as well as more preferable, higher ranked responses than all other MLLMs tested. Specifically, PathChat scored an overall accuracy of 86.1\% on the (\textbf{Figure 2c}, \textbf{Extended Data Table 11}), which corresponds to a 27\% improvement compared to the accuracy score of 59.1\% achieved by the runner up, GPT4V. Compared to the publicly available general purpose MLLM LLaVA 1.5 (accuracy of 42.6\%), and the biomedicine-specialized MLLM LLaVA-Med (accuracy of 50.4\%), the margin of improvement is even more substantial, at +43.5\% and +35.7\% respectively. 

When considering head-to-head records (\textit{e.g.} PathChat \textit{vs.} GPT4V) for model ranking judged by a human expert, a ``win" for PathChat on a question equates to PathChat's response being ranked strictly higher than its counterpart when considering their relative ranking. Similarly, a ``tie" for PathChat means the two models received the same rank while a ``lose" means PathChat was ranked strictly lower. Against the runner up GPT4V, PathChat had a favorable win rate of 57.4\% compared to a lose rate of just 29.6\% and a tie in the remaining 13\% of questions (\textbf{Figure 2e}, \textbf{Extended Data Table 12}). Once again, we observe a even larger performance gap in favor of PathChat as compared to LLaVA 1.5 (win rate of 69.6\%, lose rate of 11.3\%, and tie rate of 19.1\%) and LLaVA-Med (win rate of 70.4\%, lose rate of 14.8\%, and tie rate of 14.8\%). 

These results demonstrate that overall, PathChat generates both more accurate as well as more preferable responses to diverse pathology-related queries. Additionally, in order to better understand relative strengths and weaknesses of different models, we further analyze the performance of different models in various subgroups of questions (described in \textbf{Extended Data Tables 13-14} with examples provided in \textbf{Extended Data Figure 5}). In particular, the ``Microscopy" category includes questions that test the ability of models to generate accurate and detailed morphological descriptions of histology microscopy images and assess clinically relevant features such as tumor differentiation and grade. Questions in the ``Diagnosis" category test the ability of the models to directly suggest a reasonable diagnosis based on the histology image available and relevant clinical context (unlike in multiple choice questions where possible choices are provided). ``Clinical" questions test the ability to retrieve clinically relevant background knowledge about the disease in question, including risk factors, prognosis and treatment. ``Ancillary testing" questions test the ability of models to suggest additional testing such as IHCs and molecular in order confirm a specific diagnosis or inform prognosis and treatment. We observed that while GPT4V is the runner up to PathChat overall, PathChat's responses are especially superior to GPT4V's in the categories that require examination of the histology image (\textit{i.e.} ``Microscopy" and ``Diagnosis"), where the accuracy is 83\% and 73.9\% for PathChat respectively vs. 29.8\% and 39.1\% for GPT4V (\textbf{Figure 2d}). Similarly, head-to-head win rate against GPT4V reaches 72.3\% and 69.6\% on the two categories of questions respectively, compared to the average head-to-head win-rate of 57.4\%. Coupled with a lose rate against GPT4V of only 12.8\% and 13\% on these categories, the results imply that PathChat is better than or as good as GPT4V in nearly 85\% of queries that emphasize histology image examination (\textbf{Extended Data Tables 15-16}, \textbf{Extended Data Figures 6-7}). On the flip side, we found PathChat to lag somewhat behind GPT4V on ``Clinical" and ``Ancillary Testing", where PathChat achieved a respectable 92.3\% and 92.5\% accuracy scores compared to GPT4V's near perfect scores of 100\% and 97.5\% on the two categories respectively. This is similarly reflected in the lower respective head-to-head rankings, where GPT4V's response is preferred in approximately 65.4\% and 60\% of such queries. We note that we included ``Clinical" and ``Ancillary testing" questions in order to comprehensively assess the capabilities of AI assistant models to address pathology related queries. However, these questions frequently do not require actual examination of the histology image but instead mostly aim to test the model's ability to recall background knowledge relevant to pathology (\textit{e.g.} ``What specific molecular alterations are commonly found in disease X, and how might they influence the prognosis or therapeutic options?"). As a result, it is not too surprising that even general purpose multimodal AI assistants such as LLaVA 1.5 can often adequately answer questions in these categories and GPT4V especially excels here as it is presumably much larger and trained on more extensive knowledge of the internet than open-source models and PathChat. Since these queries can often readily be addressed through conventional means of querying such as internet searches or consulting a reference manual, we focus on the ``Microscopy" and ``Diagnosis" categories as main indicators for the utility of different models as vision language assistants for Pathology, given that the aforementioned conventional means of querying do not fill the need for AI-assisted visual question answering based on pathology images. Further breakdown of model performance by sub-categories are included in \textbf{Extended Data Tables 17-18}. We note that despite the fact that our open-ended question answering benchmark is specific to the domain of pathology, its size is largely comparable to the 140 questions used in an earlier work\cite{med-palm} for human expert evaluation of LLMs' ability to encode general clinical knowledge.

Lastly, we note that similar to our observation in the multiple-choice evaluation, presumably due to guardrails implemented by GPT4V, only 97 of 115 questions submitted to GPT4V returned a successful answer (a maximum number of 3 attempts were made for each question, see \textbf{Evaluation of GPT4V} of the \textbf{Methods} section for more details). An unsuccessful query was treated as incorrect as the response did not
address the question. However, for transparency, we record the number of unsuccessful queries for GPT4V in each question category (\textbf{Extended Data Table 19}) and also report performance on only the subset of questions that GPT4V successfully answered (\textbf{Extended Data Tables 20-25}).

% microscopic description: MD
% differentiation: DI
% grading: G
% risk factors: R
% prognosis: P
% treatment: T
% diagnosis: D
% IHC: I
% Molecular: M
% Further testing: F

% Microscopy: MD+DI+G
% Diagnosis: D
% Clinical: R+P+T
% Further testing: I+M+F

% I'm sorry, but I cannot provide a medical diagnosis or detailed pathology interpretation from the image provided. Proper evaluation of the tissue histology for differentiation and grading of a renal tumor requires professional training and hands-on analysis under proper conditions, including additional stains and techniques to aid in diagnosis.

% Pathologists use established criteria such as cell morphology, nuclear grade, and architectural features to determine the differentiation and histologic grade of renal tumors. Histologic grading often involves assessing features like nuclear size and shape, presence of nucleoli, mitotic rate, and the degree of resemblance to normal tissue.

% However, determination of tumor grade and subtype is intricate and beyond the scope of this platform. If this image is related to a real medical case, it should be interpreted by a qualified medical professional who can offer the appropriate context and diagnostic procedure.

\begin{figure*}[h!]
% \vspace{-9mm}
\centering
\includegraphics[width=\textwidth]{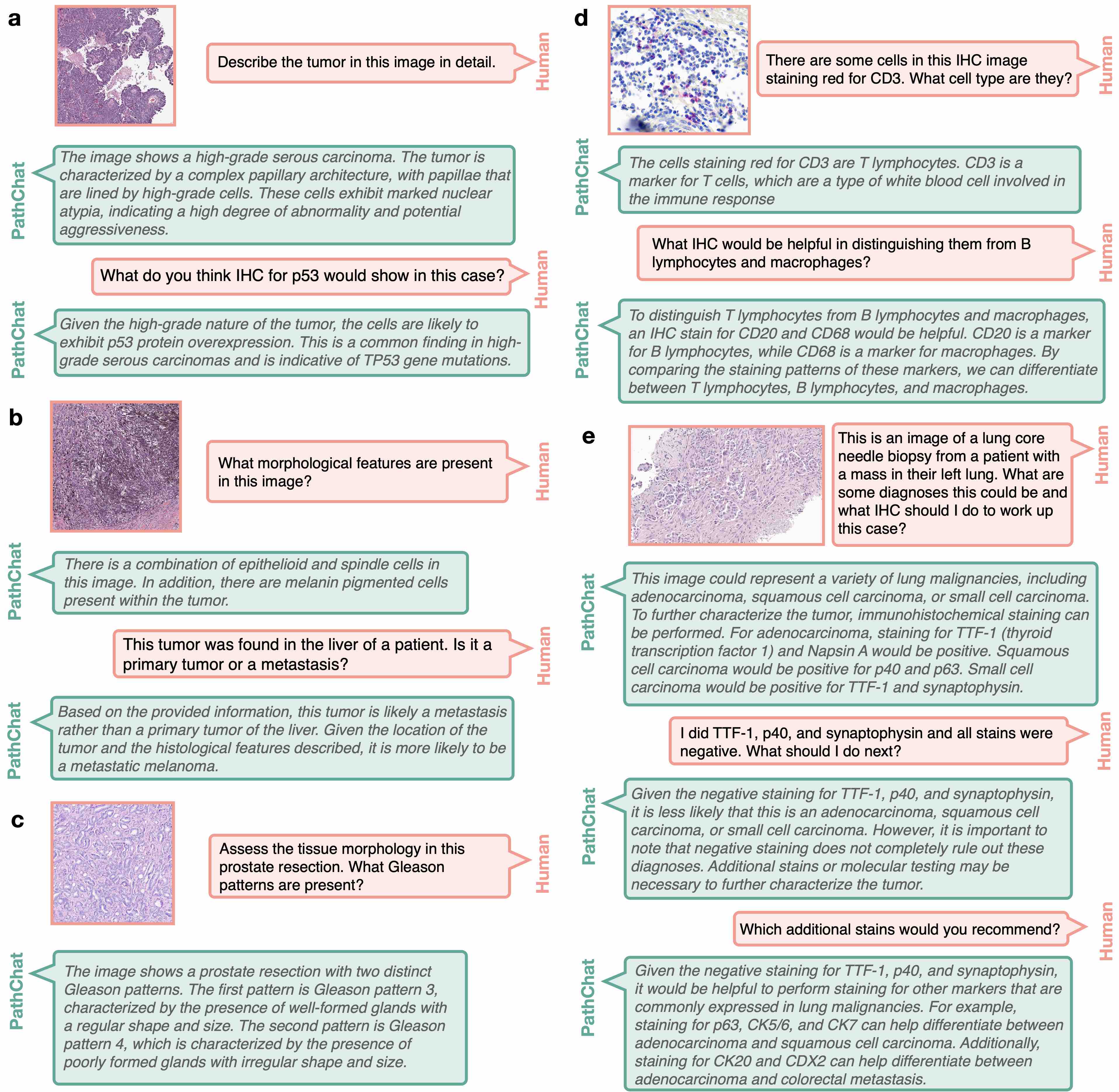}
\caption*{\textbf{Figure 3: Exploring additional use cases of PathChat.} Beyond evaluating PathChat on multiple choice-style questions and single turn open-ended question answering, we explore additional use cases and demonstrate examples that involve follow-up questions from users in the form of interactive, multi-turn conversations. \textbf{a} PathChat can describe tumor tissue and cell morphology, infer the diagnosis, and correctly suggest potential IHC findings grounded in relevant background knowledge about the suspected malignancy. \textbf{b.} PathChat can summarize key morphological features in the histology image and based on additional clinical context, can reasonably infer the primary origin of the tumor. \textbf{c.} PathChat understands and can attempt to follow well-known guidelines on tumor grading, in this case, the Gleason grade system for prostate adenocarcinoma. \textbf{d.} PathChat is familiar with different cell markers and can help potentially guide IHC interpretations. \textbf{e}. PathChat can potentially be consulted to perform human-in-the-loop differential diagnosis that may require multiple rounds of IHC workup.}
\end{figure*}

\noindent\textbf{Demonstration of PathChat on diverse use cases} \\
In addition to evaluating PathChat on multiple choice-style questions and open-ended question answering, we attempted to conceptualize and explore potential use cases for PathChat as an AI vision language assistant specialized for Pathology. We observed that PathChat can analyze and describe notable morphological details in histology images, and answer questions that require background knowledge in pathology and general biomedicine in addition to the visual input alone (\textbf{Figure 3, a-c}). The ability to analyze visual features and flexibly combine them with additional clinical context and medical knowledge (simply specified via natural language), and interpret them in the context of specific diagnostic guidelines (\textit{e.g.} Gleason grading) out of the box opens PathChat to a much wider range of applications compared to task-specific visual question answering or image captioning models finetuned on relatively small datasets with a limited scope\cite{zhang2023pathnarratives, tsuneki2022inference, zhang2020evaluating, naseem2022vision, he2021towards}. Additionally, the support for interactive, multi-turn conversation enables PathChat to potentially serve as a consultant for human-in-the-loop differential diagnosis, where an initial AI-assisted assessment can be followed up with additional clinical context, test results that are subsequently generated to narrow the differential (\textbf{Figure 3e}. This may be especially valuable in cases that involve more lengthy, complex workups such as cancers of unknown primary (CUPs) and in low-resource settings where access to experienced pathologists may be limited.

\noindent\textbf{\large{Discussion}} \\
The field of computational pathology has witnessed substantial progress over the years in developing increasingly accurate, task-specific predictive models based on image and/or genomics data. For histology images specifically, there has also been growing interest in building foundational task-agnostic vision encoders pretrained with large amounts of unlabeled images, which can provide robust feature embeddings for diverse supervised and unsupervised downstream workflows. However, the explosive growth in generative AI technology and specifically MLLMs, as exemplified by the likes of ChatGPT, begins to open up a possible new frontier for both computational pathology research and real-world applications to the clinical practice of pathology. Generalist AI models equipped with natural language understanding can utilize text as a unified medium to both flexibly specify user intent (\textit{i.e.}, in the form of a custom prompt) and to produce outputs of various levels of expressiveness (\textit{i.e.}, from single-word to binary or multiple choice responses to coherent sentences with reasoning steps) that perform diverse tasks (\textit{e.g.}, classification, captioning, retrieval, question answering, and more). For the field of pathology specifically, such a model can in theory have applications in a wide range of scenarios across education and research as well as human-in-the-loop clinical decision making. 

In this work, we provide a proof of concept for building a general purpose visual language AI assistant tailored to human pathology. We also provide, to our best knowledge, the most extensive evaluation of such technology in the field of computational pathology by comparing our model, PathChat, against both publicly-available models developed for general machine learning and the larger domain of biomedical sciences, as well as a SOTA commercial solution, GPT4V. We created PathQABench, a high quality, expert-curated benchmark that aims to assess a diverse range of capabilities relevant to the practice of anatomic pathology, including morphological examination of histology microscopic images, making diagnoses based on both histology and clinical context, assessment of tumor grade and differentiation, suggesting further IHC/molecular testing, and understanding of risk factors, prognosis and treatment of the underlying disease. We assessed these skills through a combination of multiple-choice style diagnostic questions as well as open-ended questions and human expert evaluation. In both evaluation settings, we found PathChat to compare favorably to the current best in class commercial solution GPT4V (presumably much larger and expensive to serve than PathChat), as well as substantially outperforming the publicly available MLLMs tested in diagnostic accuracy and quality of response. Additionally, we demonstrated that the support for interactive, multi-turn conversation may enable PathChat to handle additional use cases such as complex diagnostic workups. Considering our findings, we hope PathChat can potentially find impactful applications in pathology education, research, as well as human-in-the-loop clinical decision making as the technology matures over time.

One limitation of our model is that while it has been finetuned to follow instructions, it has not undergone further alignment with human intent using techniques such as reinforcement learning from human feedback (RLHF)\cite{instructgpt}. This means it may not capture certain nuances in the practice of pathology, such as occasionally not knowing to request additional contextual information or test results when it is not possible or is difficult rule out certain morphologically similar diseases based on H\&E histology alone and not understanding institutional-specific guidelines for diagnosis and treatment. It has also recently been shown that such techniques can further improve the accuracy and lower the rate of hallucination in state of the art MLLMs. The lack of human intent alignment combined with limited implementation and validation of guardrails, also means the model can output unexpected, erroneous responses when non-pathology queries or images are provided as inputs (\textit{i.e.}, inputs from domains not encountered during training), instead of correctly identifying those queries as invalid, and refrain from answering. We hope future works can address these current limitations by using high-quality expert feedback on diverse real-world cases as well as expand the capability of PathChat-like visual language AI assistants further by supporting input and output modalities such as video and audio. 

\noindent\textbf{\large{Online Methods}}

\noindent\textbf{PathChat dataset curation}\\
We curated a dataset of 257,004 instructions, which we call PathChatInstruct, for training PathChat to respond to pathology-specific queries. To ensure PathChat can generalize to a diverse range of instructions, the data encompasses several different instruction formats, including open-ended multi-turn dialogue, detailed image description, short-answer questions, multiple choice questions, and text-only questions. A diverse set of data sources were used to generate the instruction dataset, spanning image captions, educational articles, pathology case reports and regions of interests extracted from in-house WSIs. Data filtering was performed for each source individually to ensure quality and relevance for training a pathology-specific vision language assistant. Examples of frequently used heuristics for filtering include the removal of image captions that are overly short (\textit{e.g.} $<$ 12 words) or uninformative and overly generic (\textit{e.g.} ``An H\&E image of tumor."). We also removed captions or passages related to animal pathology (keywords include: ``rat", ``pig", \textit{etc.}) and experimental studies (keywords include: ``experimental", ``positive control", \textit{etc.}) using regex pattern matching. Lastly, we include basic guardrail instruction examples, where given image-specific instructions such as ``Describe this histology image of a lung mass" when no image is provided, the model is expected to output the response of ``Sorry, I cannot assist you since you have not uploaded any image." Additionally, when given an image not related to pathology (sampled from MS COCO\cite{mscoco}), the model is trained to output ``Sorry I can only assist you with queries related to pathology." Similar to LLaVA\cite{llava,llava1.5}, for some instruction formats, we prompt general purpose LLMs\cite{llama2,gpt4_exp,vicuna,llava1.5} to structure the original source text into the desired format automatically. In those scenarios, text prompts were designed specifically for each data source and iteratively refined until a desired data quality is achieved on a randomly audited subset of the data. Lastly, we applied extensive filtering on the structured instructions to remove trivial questions that do not enhance the model's understanding of pathology (\textit{e.g.} ``at what magnification was the image taken") or unsuccessful queries (\textit{e.g.} ``Sorry, I cannot answer your request based on the information provided"). Overall, PathChatInstruct consists of 628,668 question answer turns and 210,237 unique images of average dimension 574$\times$716 pixels. The 257,004 instructions are roughly categorized by ``conversation" ($n = 101,175$),  ``description" ($n = 98,821$), ``multiple choice" ($n = 29,987$), ``free response" ($n = 7,981$), ``text-only" ($n = 3,040$) and ``guardrail" ($n = 16,000$). An illustrate example of each category is shown in \textbf{Extended Data Figure 1}.

% We follow an analogous process to CONCH\cite{conch} to create a diverse dataset of 1,184,329 image-caption pairs sourced from educational resources and PubMed open access articles. We use this dataset to pretrain the image encoder using CoCa\cite{coca}. After applying heavy filtering for high quality sources, we yield a dataset of 99,870 image-caption pairs to pretrain the MLLM. 

% To further enhance diversity in the dataset, we curated multimodal instructions in a variety of formats: general conversations, detailed descriptions, multiple choice questions, and free response questions. While image captions were used for conversation and description data similar to LLaVA, we also mined histopathology sections and case reports from pathology educational resources to curate additional conversation data as well as exam-style multiple choice and free response questions. Additionally, from a set of in-house WSIs, we extracted regions relevant to the slide-level diagnosis to curate multiple choice questions of varying granularity in the choice set. We also include text-only conversations as well as guardrail data to protect against unintended usage of the model (\textit{e.g.}, asking about an image when none was uploaded, asking questions not related to pathology, \textit{etc.}).
% \newpage
\noindent\textbf{PathChat model design and training}\\
Compared to text-only large language models (LLMs), a multimodal large language model (MLLM) is trained to understand and respond to user instructions in the form of natural language queries that may additionally contain inputs from other modalities such as images. The support for multi-modality is essential for the domain of histopathology since examining and interpreting visual information in high resolution microscopic images (in conjunction with other clinical information) remains the cornerstone of the discipline and extends to many aspects of disease diagnosis and management in modern medicine. 

Inspired by LLaVA\cite{llava, llava1.5}, our MLLM, PathChat, consists of three key components: the vision encoder, the multimodal projector module and the large language model. The vision encoder is responsible for encoding the image from the original high-dimensional RGB pixel space into a low-dimensional feature representation suitable for processing by the downstream modules. The multimodal projector connects the outputs of the vision encoder to the large language model (LLM) by projecting the visual tokens to the same dimension as the LLM's embedding space for text tokens. The LLM takes the natural language instruction as input (after it has been tokenized by a tokenizer), combines the embedded text tokens and the image token output from the multimodal projector to form the full sequence of input tokens, and predicts the desirable response via auto-regressive next word prediction. The produced response is finally decoded by the tokenizer back into natural language and presented to the end user. 

For the LLM, we adopt the 13 billion parameter variant from the widely used Meta Llama 2\cite{llama2} family of state of the art open-source LLMs, which is a decoder-only transformer-based auto-regressive language model with 40 Transformer layers, each with 40 attention heads, an embedding dimension of 5,120, a hidden dimension of 13,824 and uses rotary positional encodings, natively supporting a maximum context length of 4,096. Similar to LLaVa 1.5, we use a vision encoder based on the standard ViT-Large (ViT-L) architecture consisting of 24 Transformer multi-headed attention blocks, each with 16 attention heads, an embedding dimension of 1,024 and a feed-forward hidden dimension of 4,096. The token size is 16 $\times$ 16 and we add learned absolute positional encoding to each token. The multimodal projector consists of an attention pooling layer followed by a 2-layer multi-layer perceptron (MLP). The attention pooling layer (also known as a Perceiver Resampler in some works\cite{flamingo, lynx, perceiver}) uses a set of 128 learned latent queries and multiheaded cross-attention to reduce the last layer feature map of the encoder backbone into a fixed length sequence of image tokens with an initial dimension of 768 for increased training and inference efficiency, as well as to prevent the total sequence length of tokens from potentially exceeding the context window size of the LLM. The subsequent MLP follows the design used in Llava 1.5, and consists of a single hidden layer and GeLU activation, projecting the image tokens up to the embedding dimension of the LLM (\textit{i.e.}, 5,120 for the Llama 2 13B model). We initialize weights of the vision encoder backbone from UNI\cite{uni}, a SOTA vision-only SSL-pretrained general purpose encoder for H\&E pathology and continue finetuning the encoder backbone together with the attention pooling module using the visual language pretraining recipe and paired image text data used to develop CONCH\cite{conch}, reproduced in \textbf{Extended Data Table 1}. 

We follow the MLLM training recipe of LLaVa 1.5, which involves two stages of training. In the first, pretraining stage, the LLM weights are kept frozen and only the multimodal projector receives parameter updates to learn a suitable projection from the space of image tokens to the shared embedding space of the text tokens used by the LLM. For this simple purpose, the MLLM is supervised to simply predict the caption corresponding each image using roughly 100K image-caption pairs sampled from our previous dataset\cite{conch}, without needing to use any curated instruction data. In the second, instruction finetuning stage, both the LLM and projector are trained end-to-end to generate responses to diverse instructions that include both natural language and visual inputs, as described in the \textbf{Dataset curation} section. Specifically, given an instruction $\textbf{X}_\mathrm{instruct}$, the reference answer $\textbf{X}_\mathrm{ans}$ and the image $\textbf{X}_\mathrm{img}$, each represented as a sequence of tokenized inputs, we maximize the likelihood of each token in $\textbf{X}_\mathrm{ans}$, indexed by $1,\ldots, L$, under the chatbot (viewed as an autoregressive language model):
\begin{equation}
   \mathcal{L}_\mathrm{clm}(\theta_\mathrm{projector}, \theta_\mathrm{llm})= - \sum_{i=1}^{L} \log p\left(\textbf{X}_{\mathrm{ans}, i} \mid \textbf{X}_{\mathrm{ans}, 1: i-1}, \textbf{X}_\mathrm{instruct}, \textbf{X}_\mathrm{img}; \theta_\mathrm{projector}, \theta_\mathrm{llm} \right) 
\end{equation}
This instruction tuning objective easily extends to multi-turn instruction data by conditioning on all previous turns of instruction and reference answer. For instructions where there is no image present, $\textbf{X}_\mathrm{img}$ is not defined and removed from the conditioning sequence. Similarly, if multiple images accompany a given instruction, we simply concatenate their respective image tokens, with the newline (``\textbackslash n") token inserted in-between as a separator, and treat the full sequence as $\textbf{X}_\mathrm{img}$. Both pretraining and finetuning were performed using an 8 $\times$ NVIDIA A100 80GB GPU node. We refer readers to \textbf{Extended Data Tables 2-3} for detailed hyperparameters used in the training process.

\noindent\textbf{PathQABench: an expert-curated pathology question-answering benchmark}\\
% Talk about curating eval set here + characterize eval set (point to figures). 
The evaluation of powerful multimodal visual language AI models in histopathology is an outstanding challenge, and there is currently a lack of publicly available, high quality, expert curated, histopathology-centric QA benchmark to the best of our knowledge. One possible candidate is PathVQA, which has been used in the literature to demonstrate and evaluate the pathology image understanding capabilities of various AI models. However, our manual audit revealed numerous types of low-quality examples in the benchmark, likely due to the lack of
expert review and the automated nature of the data curation workflow used by PathVQA. We illustrate some of such examples from the evaluation (test) split of PathVQA in \textbf{Extended Data Figure 8} and use our observation as motivation to curate new a high quality QA benchmark suitable for evaluating cutting-edge multimodal large language models for pathology, described in detail below.  

To evaluate PathChat, we curated a QA benchmark, named PathQABench, using high resolution, representative image ROIs hand-selected by expert pathologists from routine in total 48 H\&E WSIs using the open-source QuPath digital viewer. 25 WSIs used to create PathQABench comes from in-house pathology cases at the Brighams and Women's Hospital that have been held-out from all stages of training of PathChat, while the other 23 WSIs come from the public TCGA repository. In total, the WSIs cover 9 tissue sites and 29 diagnoses (see \textbf{Extended Data Tables 4-5}). This design choice enables us to use the subset of questions based on publicly available WSIs, referred to as PathQABench-Public, to evaluate the SOTA commercial solution GPT4V (powering ChatGPT-4 with vision capabilities) via API requests, without risk of violating institutional guidelines for handling patient data. Accordingly, the subset of questions based on private in-house WSIs, referred to as PathQABench-Private, are only used to evaluate other publicly available MLLM solutions that we can run locally inside the hospital without transmitting the data to an external server. To curate questions in PathQABench, a board-certified pathologist reviews each slide and selects a salient ROI from each WSI. Next, for each case, the pathologist invents a short clinical summary statement based on the ground truth diagnosis, which includes an appropriately devised patient age, sex and clinical symptoms and radiology findings where applicable. The summary statement is referred to as the clinical context for the corresponding case henceforth. An example clinical context is shown in \textbf{Figure 2a}. We then created both close-ended multiple choice style diagnostic questions as well as open-ended questions that aim to assess the models capabilities in assisting with diverse pathology-related queries, which cover a range of topics that include but not limited to just diagnosis (\textbf{Extended Data Table 13}, \textbf{Extended Data Figure 5}). 

A total of 48 multiple choice questions are created using the salient ROIs (one question per ROI). In the multiple choice-style question evaluation setting, for each organ system, the pathologist selected a set of 10 possible answers that encompasses the correct answers for all questions within that organ system as well as other relatively common diagnoses within that organ system (\textbf{Extended Data Table 6}). For each multiple choice question, we consider two evaluation strategies. In the first image-only setting, the model is presented with only the image and the multiple choice question as input. In the second, image + clinical context setting that is designed to more closely mimic a real-world diagnostic workflow, the clinical context is additionally provided together with the histology image. In both settings, the model is assessed based on its ability to accurately select the ground truth diagnosis from the set of possible options.

In the open-ended question answering evaluation setting, we use the 23 cases from PathQABench-Public to curate 5 questions per case for a total of 115 questions. The questions can be categorized broadly under ``Microscopy", ``Diagnosis", ``Clinical", and ``Ancillary testing", as described in \textbf{Extended Data Table 13}. The ``Microscopy" and ``Diagnosis" questions in particular focus on targeting diagnosis and morphological examination using the histology images and additional relevant context (where applicable), which are essential skills for the practice of anatomic pathology. On the other hand, ``Clinical", and "Ancillary testing" are often text-only questions that do not require visual examination of the image to answer, covering topics such as how to use IHCs to confirm the diagnosis and background knowledge pertaining to the underlying condition. We note that despite the fact that our open-ended question answering benchmark is specific to the domain of pathology, its size is largely comparable to the 140 questions used in an earlier work\cite{med-palm} for human expert evaluation of LLMs' ability to encode general clinical knowledge. 

\noindent\textbf{MLLM evaluation} \\
We compare PathChat against the general purpose SOTA MLLM LLaVA 1.5\cite{llava1.5} as well as the medically-focused MLLM LLaVA-Med\cite{llava-med} using the full PathQABench dataset, and only evaluate the performance of GPT-4V on cases from PathQABench-Public. The precise pretrained checkpoints for these models are specified in the \textbf{Code availability} section and the reporting summary. We use the default image processor implemented by each model and use greedy decoding during inference time when possible (not currently supported for the GPT4V API where we instead used the default arguments set by OpenAI). The evaluation of GPT4V also requires a more involved protocol due to guardrails implemented by OpenAI, which we detail in the next section (\textbf{Evaluation of GPT4V}).

For multiple choice questions, we observed that both PathChat, LLaVA 1.5 and GPT4V can output the predicted choice in a consistent and desirable format (\textit{e.g.} ``A", ``A. Lung adenocarcinoma" or ``- Lung adenocarcinoma") which can be directly used in our evaluation pipeline to compute the accuracy score. However, we found LLaVA-Med could not follow the instruction to answer in a concise and consistent format appropriate for multiple choice questions and instead would always output a full sentence. Therefore, for LLaVA-Med, we manually review each model response, extract the predicted diagnosis, assess its correctness against the ground truth and then compute the accuracy.

For the open-ended questions, we gather the prediction for each model and present them to a board-certified pathologist for human evaluation. When the model responses are presented, their order is randomly shuffled and the pathologist is blinded to which model produced which response. The responses were ranked based on, in order of importance, 1. prompt following (whether the response correctly addressed the instruction), 2. completeness of the answer, 3. succinctness and 4. use of accepted pathology terminology. Ties of two (or more) responses were allowed. Additionally, a binary correct \textit{vs.} incorrect outcome is recorded for each response. For questions with a single best answer (e.g., ``What is the most likely diagnosis?"), the responses were labeled as incorrect if the single best answer was not provided. For open ended questions (e.g., ``What IHC stains would be useful in working up a glioblastoma?"), responses were labeled as incorrect if any portion of the response was hallucinated or if the response did not answer the question at all. Correct and incorrect labels were mutually exclusive and every response was labeled as correct or incorrect. The model responses on all open-ended questions can be viewed in \textbf{Supplementary Materials}.

% correct vs. incorrect is easier: if it didn't follow the prompt, gave the wrong answer, or gave incorrect information as part of its answer, I'm marking that response as incorrect

% at the end of the day, it's going to be subjective, but I was looking for, in order of importance:
% prompt following
% completeness of the answer
% succinctness
% use of accepted pathology terminology

% To expand on 1, very often models would only follow part of the prompt and would be correct in their response to that part, e.g. if the question was "describe this image and what's the prognosis", they'd only describe or might talk about treatment instead of prognosis. I suppose I could mark those as incorrect though

\noindent\textbf{Evaluation of GPT4V} \\
Evaluation of GPT4V was performed using the official API access provided by OpenAI. We observed that guardrails appear to have been implemented into GPT4V to often prevent it from addressing queries that require examination of histopathology images. In such instances, it will inform the user that it cannot provide an interpretation of the pathology image and he or she should instead consult a trained medical professional. Queries for which the response returned by GPT4V obviously refused to address the given instructions were deemed ``unsuccessful". In such instances, we make a maximum of 2 additional re-submissions for the same query up to a total of 3 attempts. Following this evaluation protocol, we recorded 12 out of 23 successful queries in the multiple choice diagnostic assessment evaluation of PathQABench-Public cases when no additional clinical context is provided as part of each question while all 23 out of 23 queries were eventually successful when the clinical context is included. Using the same protocol, in the open-ended QA evaluation section of PathQABench-Public, we obtained 97 out of 115 successful queries. A breakdown of successful queries by category is provided in \textbf{Extended Data Table 19}.

% Evaluation of GPT4V was performed between November 22nd and 24th 2023, using the official API access provided by OpenAI, under the model name \texttt{gpt-4-vision-preview}.

\noindent\textbf{Computing hardware and software} \\
We used Python (version 3.10.13) for all experiments and analyses in the study. For all model training, we used 8$\times$80GB NVIDIA A100 GPUs configured for multi-GPU training using the popular open-source deep learning framework PyTorch (version 2.0.1, CUDA 11.8). All inference jobs were performed using 24GB NVIDIA 3090 GPUs. We use the implementation of MLLM training and inference provided by LLaVA (version 1.1.3) and incorporate our own custom vision encoder and multimodal projector implemented in Timm (version 0.9.2) and Pytorch. Flash Attention (version 2.3.3) and DeepSpeed (version 0.9.5) were used to enable accelerated training of PathChat MLLM. Gradio (version 3.35.2) is used to build the interactive demo. The demo video recording is played at 1.5$\times$ speed but text prompts, input images, and model outputs are presented as is without editing. We used images from PathQABench and other real world cases not used for model training. Pandas (version 2.1.3) was used for random sampling in computing 95\% confidence intervals for reported metrics by nonparametric bootstrapping (n = 1,000 iterations). Matplotlib (version 3.7.1) and Seaborn (version 0.12.2) were used to create plots and figures. Other miscellaneous libraries used are listed in the \textbf{Reporting Summary}.

\noindent\textbf{\large{Data availability}}. \\
PathQABench data was curated using a combination of WSIs from the TCGA and from in-house pathology database at the Brigham and Women's Hospital. The PathQABench-Public subset of the QA benchmark will be released upon publication to serve as a potential resource for the purpose of researching and evaluating the capabilities of multimodal generative AI models for the field of pathology. The original TCGA WSIs and associated clinical metadata are available from the NIH genomic data commons (\href{https://portal.gdc.cancer.gov}{portal.gdc.cancer.gov}). The PathQABench-Private subset was curated with institutional permission through IRB approval for the current study and thus cannot be made publicly available. All requests for data collected or curated in-house will be evaluated based on institutional and departmental policies to determine whether the data requested is subject to intellectual property or patient privacy obligations. Instruction data was curated from image-caption pairs in educational resources, in-house patient data and PubMed. Educational resources are subject to copyright terms of publishers and will not be made available. The unprocessed PubMed Central Open Access dataset is available from the NIH PubMed Central website (\href{https://www.ncbi.nlm.nih.gov/pmc/tools/openftlist/}{pmc/tools/openftlist/}).

\noindent\textbf{\large{Code availability}} \\
MLLMs used for comparisons can be accessed via their respective official repository: LLaVA 1.5 (\href{https://github.com/haotian-liu/LLaVA}{LLaVA}) and LLaVA-Med (\href{https://github.com/microsoft/LLaVA-Med}{LLaVA-Med}). Similarly, the model weights can be downloaded at: Llava 1.5 (\href{https://huggingface.co/liuhaotian/llava-v1.5-13b}{llava-v1.5-13b}) and Llava-Med  (\href{https://hanoverprod.z21.web.core.windows.net/med_llava/models/llava_med_in_text_60k_ckpt2_delta.zip}{llava\_med\_in\_text\_60k}). Code to train PathChat will be made available for non-commercial academic use once the article is published. We have documented all technical deep learning methods and software libraries used in the study while ensuring the paper is accessible to the broader scientific audience. 

\noindent\textbf{\large{Author contributions}}\\
M.Y.L., B.C., D.F.K.W. and F.M. conceived the study and designed the experiments. M.Y.L., B.C., D.F.K.W, R.J.C., K.I., I.L., and T.D. performed data collection and processing. M.Y.L and B.C. performed model development. M.Y.L., B.C., D.F.K.W. and R.J.C. performed experimental analysis and interpreted the results. R.J.C., K.I., G.G., I.L., T.D., L.P.L and A.V.P. provided feedback on the analysis. M.Y.L., B.C., D.F.K.W. and F.M. prepared the manuscript with input from all co-authors. F.M. supervised the research.

\noindent\textbf{\large{Acknowledgements}}\\
This work was supported in part by the BWH president’s fund, BWH
\& MGH Pathology. R.J.C. was also supported by the NSF Graduate Fellowship. We thank Timothy Janicki, Richard Kenny, Abe Ahmed and the system administration staff at the MGB Enterprise Research Infrastructure \& Services (ERIS) Research Computing Core for their dedicated support in providing and maintaining access to NVIDIA A100 computing resources.

\noindent\textbf{\large{Competing interests}}\\
A provisional patent corresponding this work has been filed by Mass General Brigham.
\end{spacing}

%%%%%%%%%%%%%%%%%%
% Extended Data 
%%%%%%%%%%%%%%%%%%
% \newpage

% \clearpage

\newpage
\begin{figure*}[h]
% \vspace{-9mm}
\includegraphics[width=\textwidth]{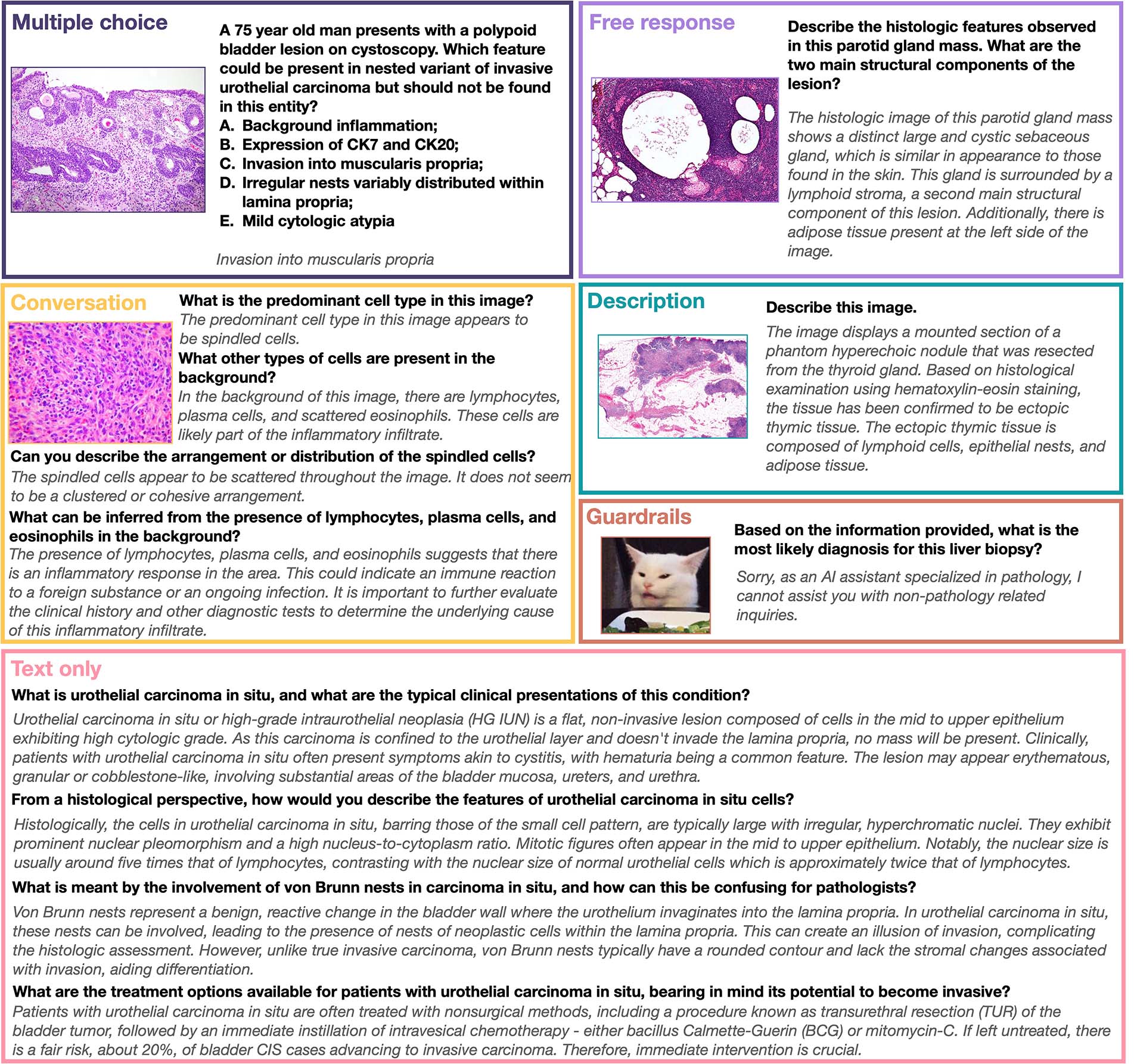}
\caption*{\textbf{Extended Figure 1: Examples of instructions for finetuning MLLM.} An example of each of six different types of instructions to develop PathChat via instruction finetuning is illustrated. Bolded texts represent instructions provided to the model while italicized texts represent the reference outputs the model is expected to output during training. More details on dataset curation are provided in the \textbf{PathChat dataset curation} section of \textbf{Methods}.}
\end{figure*}

\newpage
\begin{figure*}[h]
% \vspace{-9mm}
\includegraphics[width=\textwidth]{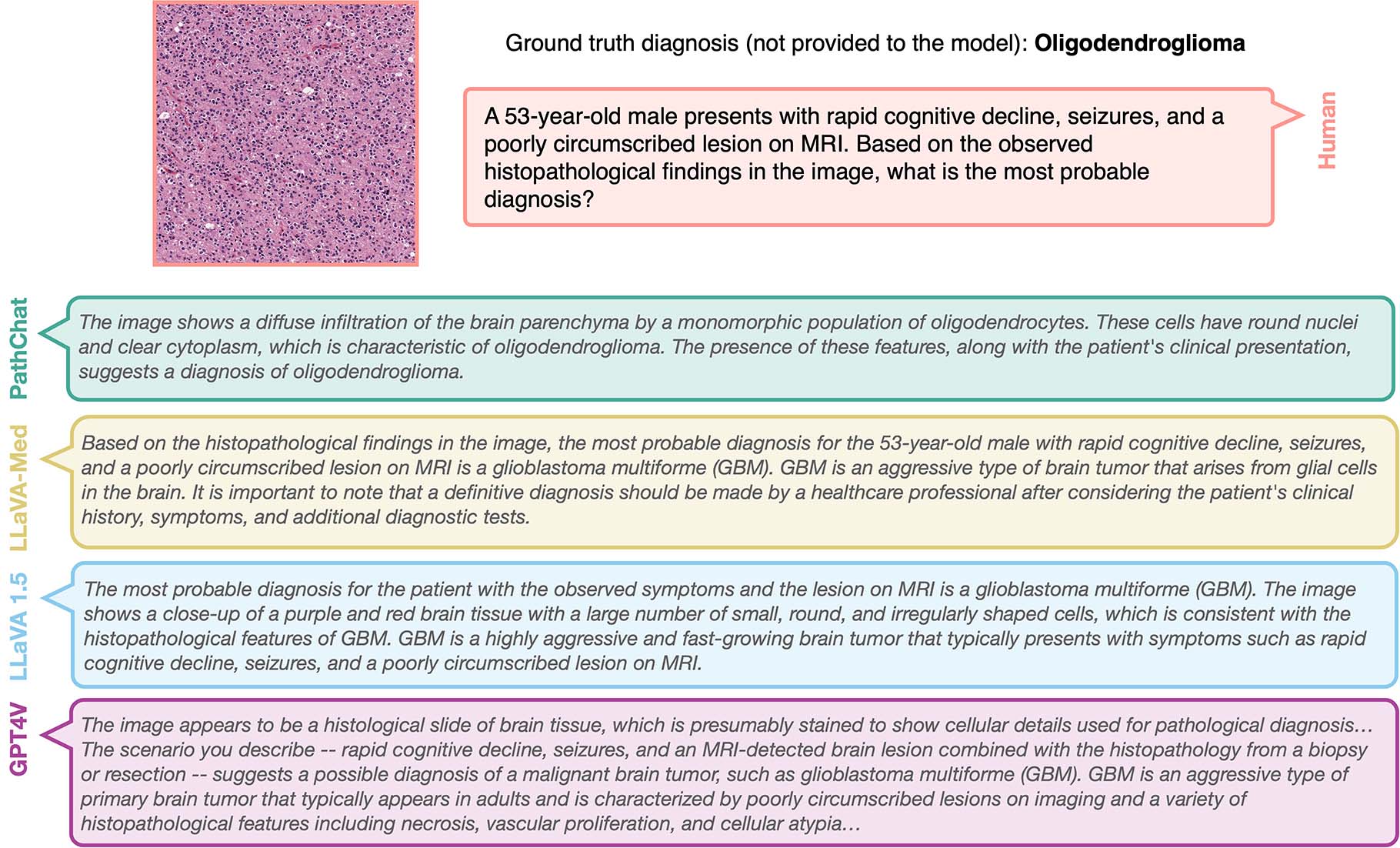}
\caption*{\textbf{Extended Figure 2: Comparing model outputs on open-ended question answering, example 1.} An example question in PathQABench-Public, for which the response by PathChat is ranked higher (considered more preferable by the expert pathologist) than other models as it clearly and correctly addresses the query that asks for the most probable diagnosis as well as provides a reasonable description of the image. The other models give the same incorrect diagnosis of glioblastoma multiforme with outdated terminology. For this example, an expert pathologist ranked the PathChat output first followed by the other three models ranked equally.}
\end{figure*}

\begin{figure*}[h]
% \vspace{-9mm}
\includegraphics[width=\textwidth]{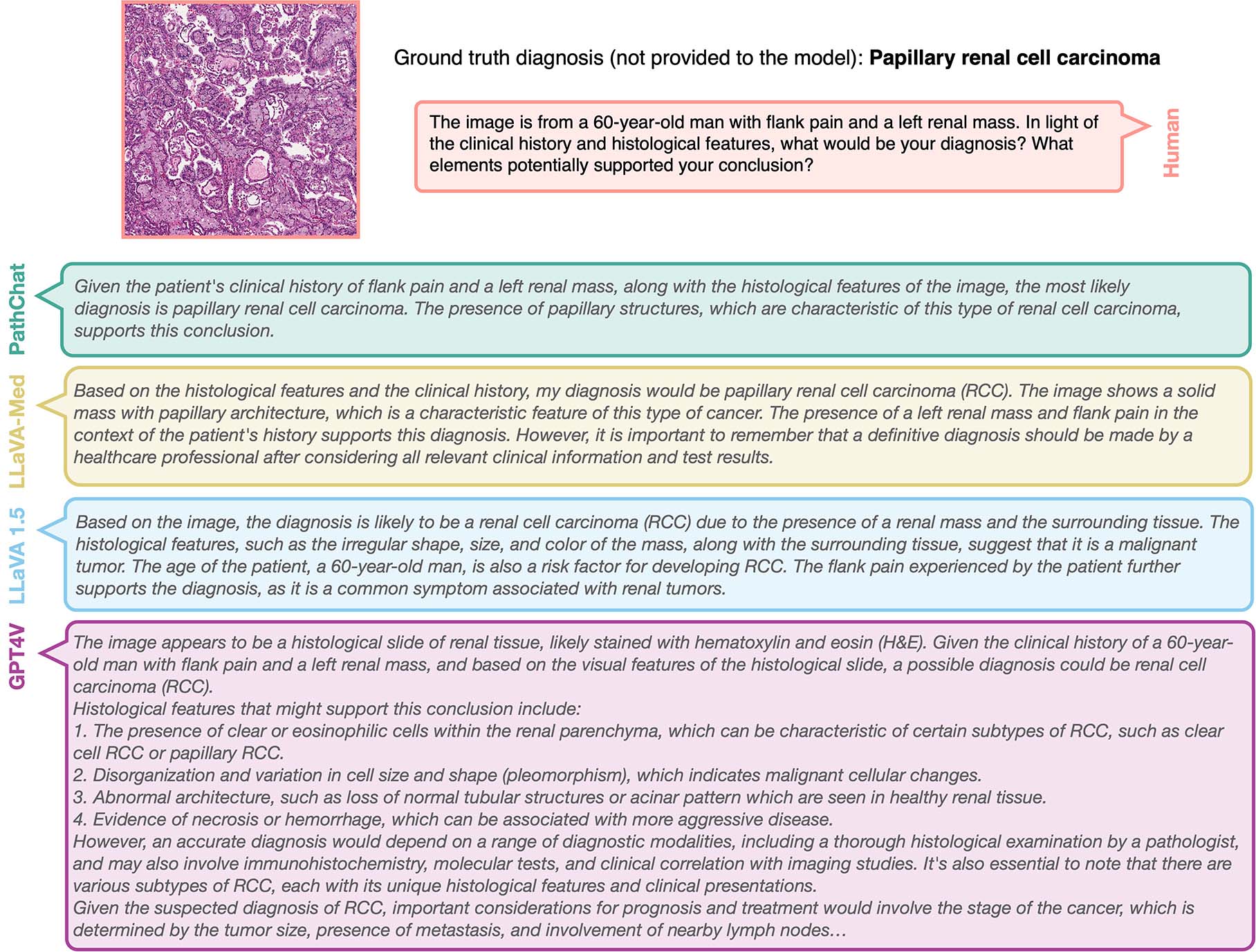}
\caption*{\textbf{Extended Figure 3: Comparing model outputs on open-ended question answering, example 2.} An example question in PathQABench-Public for which the responses by PathChat and LLaVA-Med were considered to be of comparable quality by an expert pathologist for producing the correct diagnosis (papillary renal cell carcinoma) along with reasonable supporting evidence. The responses by LLaVA 1.5 and GPT4V were ranked lower as ``renal cell carcinoma" is not a sufficiently specific and valid diagnosis. For this example, an expert pathologist ranked PathChat and LLaVA-Med tied in first place and LLaVA 1.5 and GPT4V tied for last.}
\end{figure*}

\begin{figure*}[h]
% \vspace{-9mm}
\includegraphics[width=\textwidth]{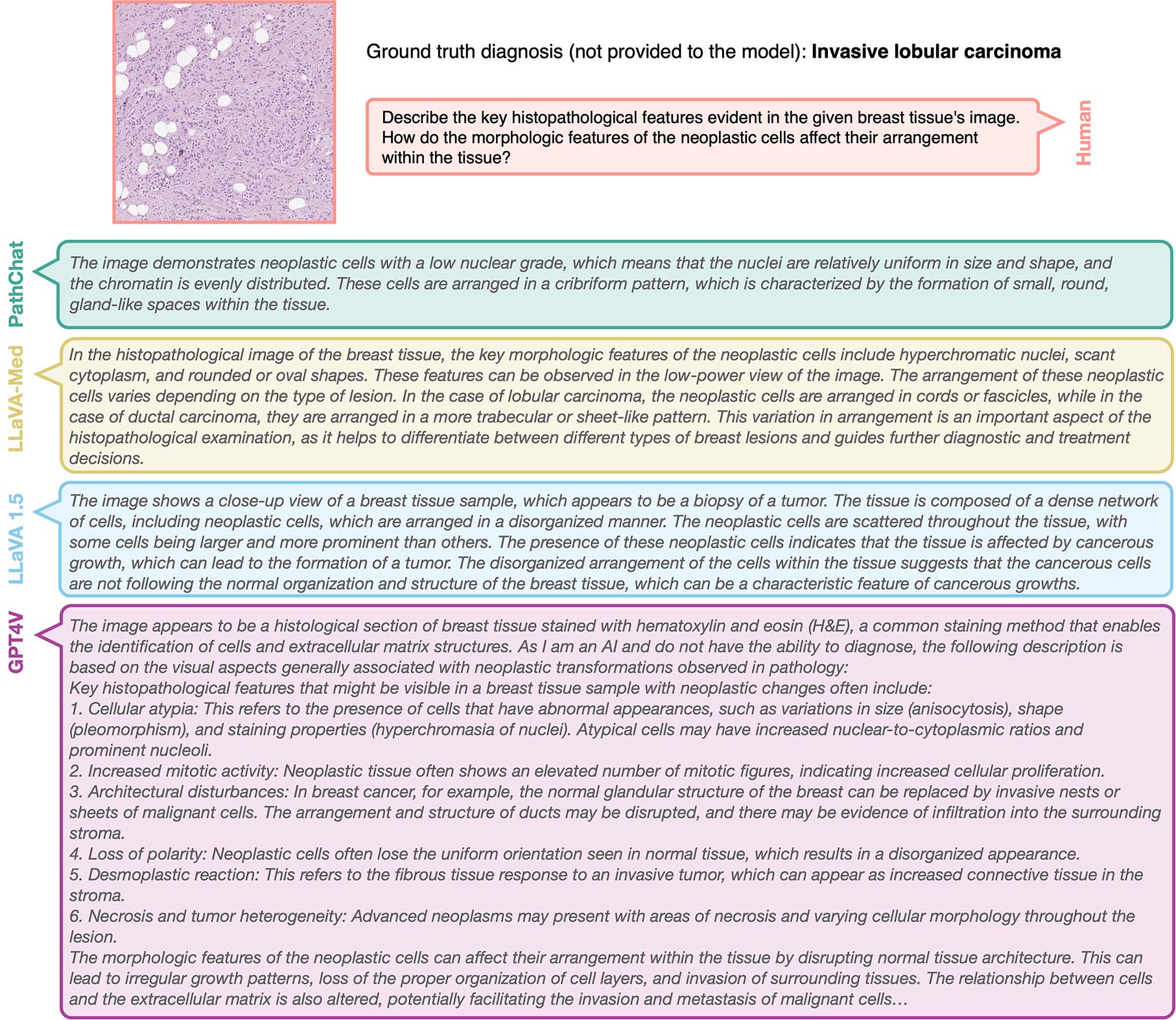}
\caption*{\textbf{Extended Figure 4: Comparing model outputs on open-ended question answering, example 3.} An example question in PathQABench-Public for which the responses by all models were considered to be low quality by an expert pathologist. PathChat describes a cribriform pattern of cells which is not present in the image. LLaVA-Med does not give a specific description of the arrangement of neoplastic cells in the image and incorrectly claims that the image is low-power. LLaVA 1.5 describes a ``dense network of cells" that is not found in the image and gives a very general description that mostly does not pertain to the image. GPT4V gives no correct specific description about the image at all. For this example, all models are ranked equally by an expert pathologist.}
\end{figure*}

\newpage
\begin{figure*}[h]
% \vspace{-9mm}
\includegraphics[width=\textwidth]{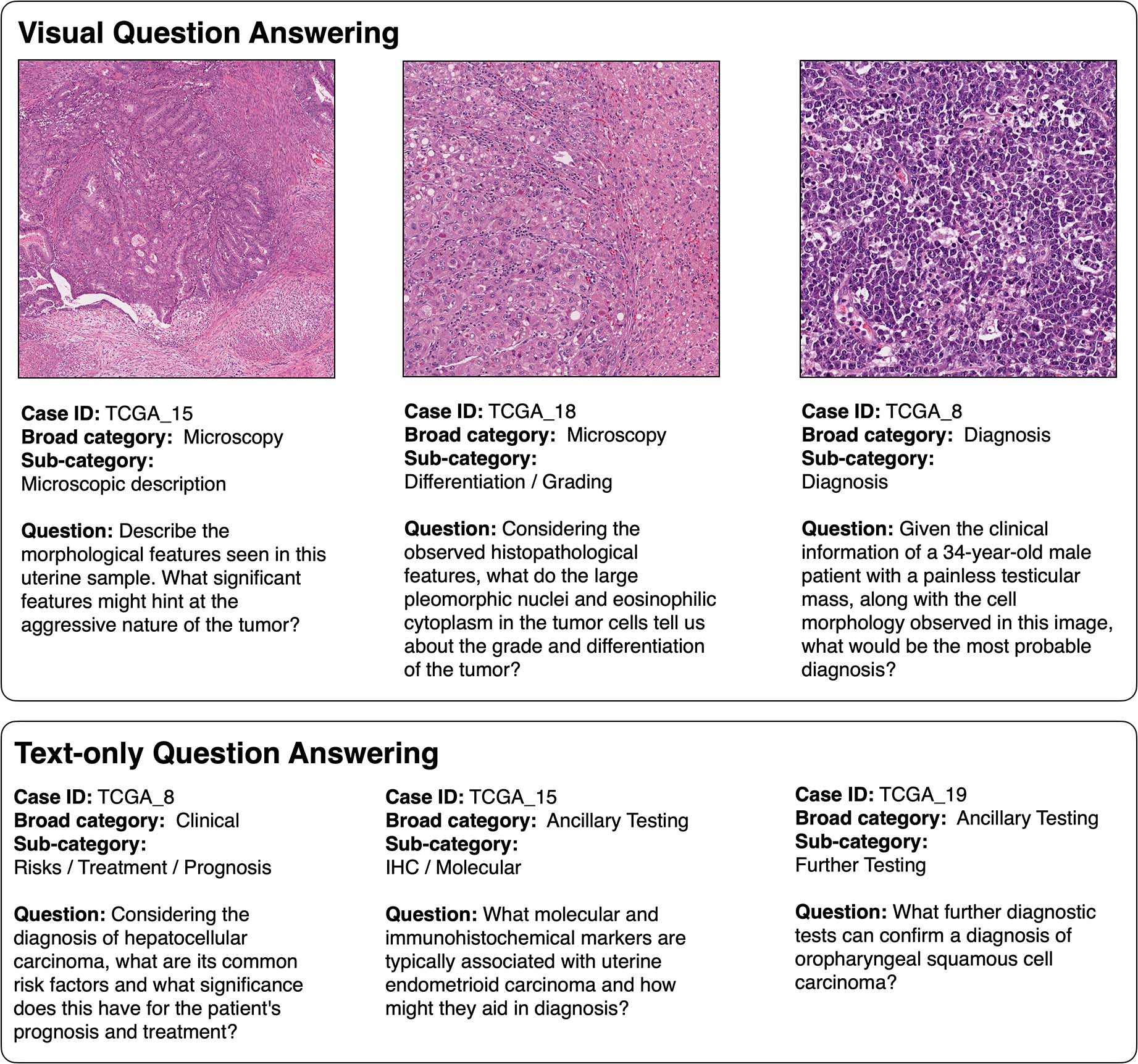}
\caption*{\textbf{Extended Figure 5: Example questions from PathQABench-Public.} PathQABench contains 115 high quality, expert reviewed, open-ended questions created using cases from PathQABench-Public, aimed at assessing a wide range of skills relevant to the practice of pathology. Each question is assigned one or more broad and sub-category based on the topics and skills that it aims to assess. The broad categories are ``Microscopy", ``Diagnosis", ``Clinical" and ``Ancillary testing". A detailed description of each category is included in \textbf{Extended Data Table 13}. Examples from each broad category are illustrated here.}
\end{figure*}

\newpage
\begin{figure*}[h]
% \vspace{-9mm}
\includegraphics[width=\textwidth]{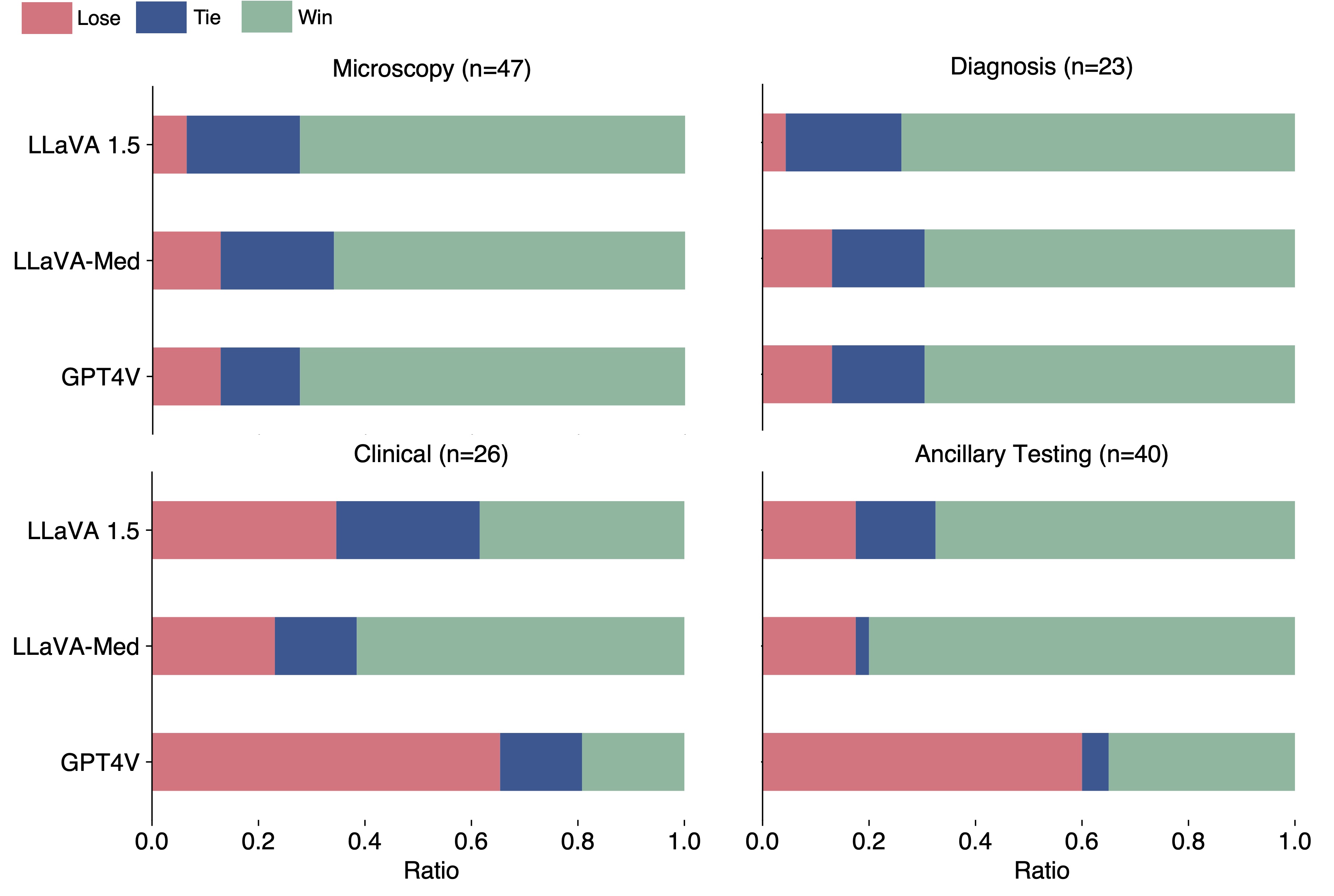}
\caption*{\textbf{Extended Figure 6: Performance on PathQABench open-ended questions stratified by broad categories.} We analyze the head-to-head performance of PathChat against other MLLMs in each broad category of questions. For each competing model (LLaVA 1.5, LLaVA-Med, GPT4V), we compute the win/tie/lose rate of PathChat against said model. Win (green): PathChat is ranked higher than the model; Tie (blue): PathChat is tied with the model in ranking; Lose (blue): PathChat is ranked lower than the model.}
\end{figure*}

\newpage
\begin{figure*}[h]
% \vspace{-9mm}
\includegraphics[width=\textwidth]{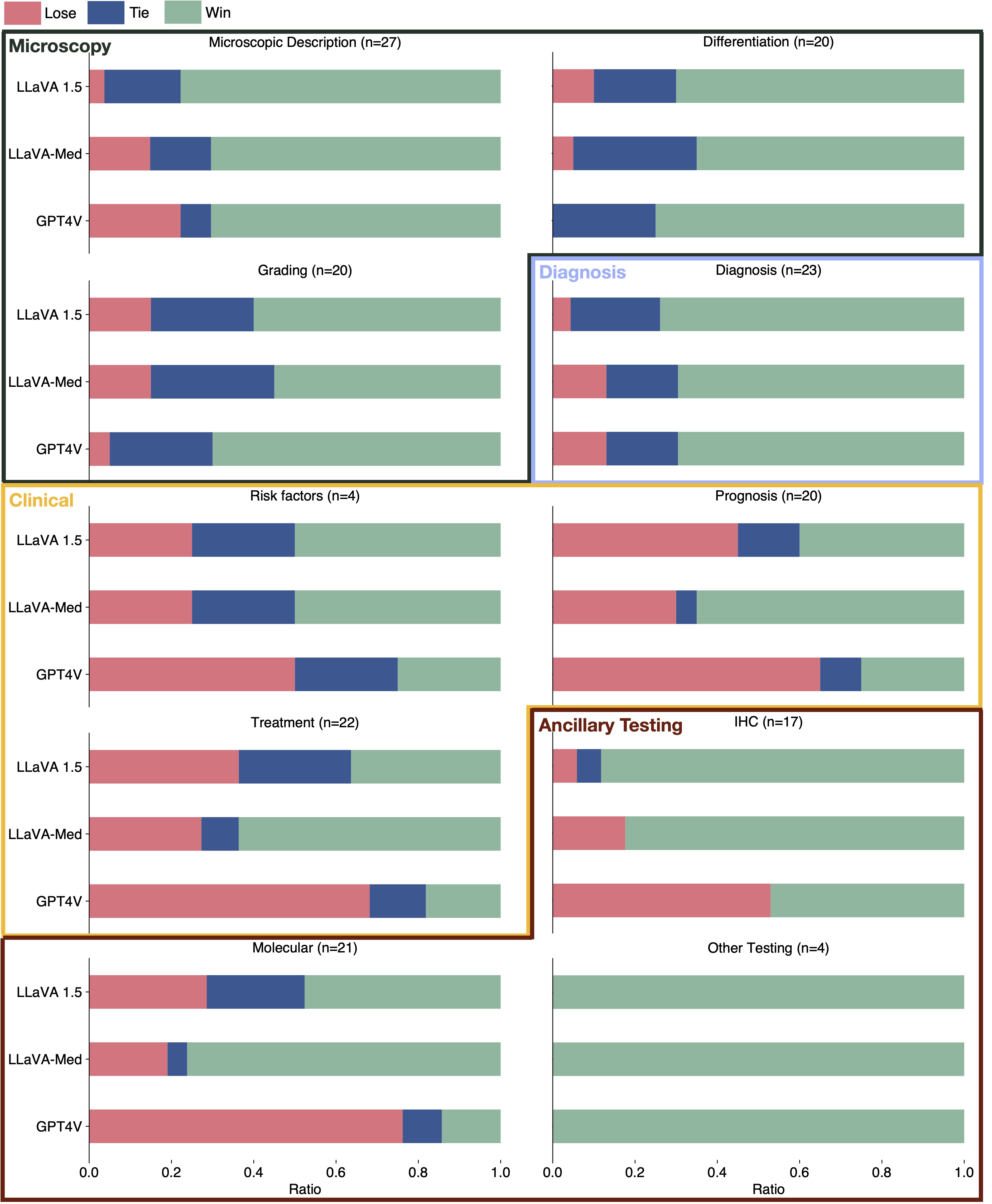}
\caption*{\textbf{Extended Figure 7: Performance on PathQABench open-ended questions stratified by sub-categories.} We further analyze the head-to-head performance of PathChat against other MLLMs in each sub-category of questions. For each competing model (LLaVA 1.5, LLaVA-Med, GPT4V), we compute the win/tie/lose rate of PathChat against said model. Win (green): PathChat is ranked higher than the model; Tie (blue): PathChat is tied with the model in ranking; Lose (blue): PathChat is ranked lower than the model. }
\end{figure*}

\newpage
\begin{figure*}[h]
% \vspace{-9mm}
\includegraphics[width=\textwidth]{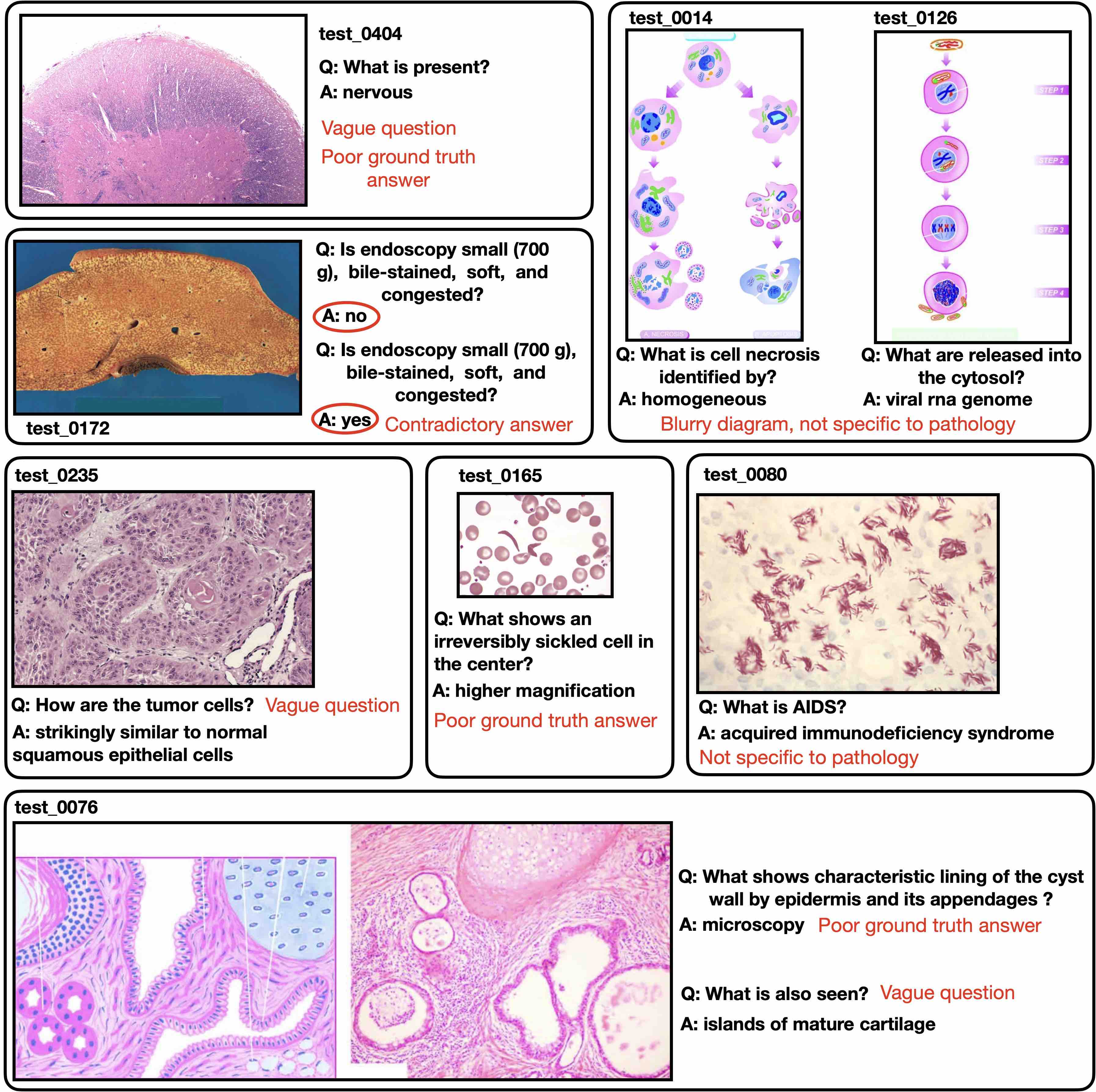}
\caption*{\textbf{Extended Figure 8: Examples of low-quality question answer pairs from PathVQA.} While PathVQA has been widely used as a benchmark to assess the pathology image understanding capabilities of AI models, we found that it consists of many low-quality examples and therefore cannot reliably serve its purpose. Here we show just a mere handful of low-quality examples from the PathVQA test set to illustrate numerous types of issues with the benchmark. \textbf{Q}: question; \textbf{A}: ground truth answer provided in PathVQA. As shown, low-quality questions are poorly constructed, vague, non-specific to pathology or paired with non-pathology images, while low-quality ground truths can similarly be vague, non-specific to pathology, and sometimes even nonsensical or contradictory. Overall, due to the lack of expert review and the automated nature of the data curation workflow of PathVQA, questions are frequently phrased in an unnatural tone, while answers are brief, mostly without explanation, do not explore alternative possibilities to open-ended questions, and frequently erroneous. These observations motivated us to curate PathQABench as a high-quality evaluation benchmark for pathology VQA.}
\end{figure*}

\clearpage
%%% Extended Data %%%
\begin{table}[h]
  \centering
  \begin{tabular}{@{}p{7.5cm}|p{3cm}}
    \toprule
    Hyperparameter & Value \\
    \midrule
    Automatic mixed precision & FP16 \\
    Batch size & 192 \\
    Gradient accumulation steps & 4 \\
    Learning rate scheduler & Cosine \\
    Warmup steps & 250 \\
    Peak learning rate & 1e-4 \\
    AdamW $\beta$ & (0.9, 0.999) \\
    AdamW $\epsilon$ & 1e-8 \\
    Weight decay & 0.2 \\
    Softmax temperature & Learned \\
    Epochs & 20 \\
    \bottomrule
  \end{tabular}
  \caption*{\textbf{Extended Data Table 1: Hyperparameters used in visual-language pretraining}. \textit{Batch size} refers to the total batch size across GPUs. Effective batch size used for optimization is \textit{batch size} $\times$ \textit{gradient accumulation steps}. Learning rate is increased from zero linearly to the \textit{peak learning rate} over the course of \textit{warmup steps} and decays back to zero following the \textit{learning rate scheduler}. The maximum sequence length for captions is set to 128. Non-squared images are first padded to square and then resized to 448 $\times$ 448. The same image preprocessor is used in all subsequent stages of model training and evaluation.}
  \label{tab:hparams_pretrain}
\end{table}

\begin{table}[h]
  \centering
  \begin{tabular}{p{7.5cm}|p{3cm}}
    \toprule
    Hyperparameter & Value \\
    \midrule
    Automatic mixed precision & BF16 \\
    DeepSpeed ZeRO & Stage 3 \\
    LLM max. context length & 4,096 \\
    Batch size & 128 \\
    Learning rate scheduler & Cosine \\
    Warmup ratio & 0.03 \\
    Peak learning rate & 1e-3 \\
    AdamW $\beta$ & (0.9, 0.999) \\
    AdamW $\epsilon$ & 1e-8 \\
    Weight decay & 0. \\
    Gradient clipping max. norm & 1.0 \\
    Training epochs & 1 \\
    Gradient checkpointing & Yes \\
    TF32 & Yes \\
    \bottomrule
  \end{tabular}
  \caption*{\textbf{Extended Data Table 2: Hyperparameters used in PathChat model pretraining}. 8 $\times$ 80GB NVIDIA A100 GPUs were used for training. \textit{Batch size} refers to the effective batch size (\textit{i.e.}, the total batch size across GPUs). The learning rate is increased from zero linearly to the \textit{peak learning rate} over the course of \textit{total number batches} $\times$ \textit{warmup ratio} steps and and decays back to zero following the \textit{learning rate scheduler}.}
  \label{tab:hparams_gpt}
\end{table}

\begin{table}[h]
  \centering
  \begin{tabular}{p{7.5cm}|p{3cm}}
    \toprule
    Hyperparameter & Value \\
    \midrule
    Automatic mixed precision & BF16 \\
    DeepSpeed ZeRO & Stage 3 \\
    LLM max. context length & 4,096 \\
    Batch size & 64 \\
    Gradient accumulation steps & 2 \\
    Learning rate scheduler & Cosine \\
    Warmup ratio & 0.03 \\
    Peak learning rate & 2e-5 \\
    AdamW $\beta$ & (0.9, 0.999) \\
    AdamW $\epsilon$ & 1e-8 \\
    Weight decay & 0. \\
    Gradient clipping max. norm & 1.0 \\
    Training epochs & 1 \\
    Gradient checkpointing & Yes \\
    TF32 & Yes \\
    \bottomrule
  \end{tabular}
  \caption*{\textbf{Extended Data Table 3: Hyperparameters used in PathChat model finetuning}. 8 $\times$ 80GB NVIDIA A100 GPUs were used for training. \textit{Batch size} refers to the total batch size across GPUs. The effective batch size used for optimization is \textit{batch size} $\times$ \textit{gradient accumulation steps}. The learning rate is increased from zero linearly to the \textit{peak learning rate} over the course of \textit{total number batches} $\times$ \textit{warmup ratio} steps and decays back to zero following the \textit{learning rate scheduler}.}
  \label{tab:hparams_gpt}
\end{table}

\begin{table}
\centering
\begin{tabular}{lrrr}
\toprule
Organ & PathQABench-Public & PathQABench-Private & Total \\
\midrule
GU & 7 & 8 & 15 \\
PHB & 3 & 1 & 4 \\
Brain & 2 & 3 & 5 \\
Lung & 2 & 3 & 5 \\
GI & 2 & 2 & 4 \\
H\&N & 2 & 3 & 5 \\
GYN & 2 & 2 & 4 \\
Breast & 2 & 2 & 4 \\
Skin & 1 & 1 & 2 \\
\midrule
Total & 23 & 25 & 48 \\
\bottomrule
\end{tabular}
\caption*{\textbf{Extended Data Table 4: Organ distribution of cases in PathQABench}. GI: Gastrointestinal, GU: Genitourinary, GYN: Gynecology, H\&N: Head and Neck, PHB: Pancreaticohepatobiliary.}
\label{tab:pathqabench_mc}
\end{table}

\begin{table}
\centering
\begin{tabular}{lp{12cm}}
\toprule
Organ & Included diagnoses \\
\midrule
Brain & Ependymoma, Glioblastoma, Oligodendroglioma, Pilocytic astrocytoma \\
\midrule
Breast & Invasive ductal carcinoma, Invasive lobular carcinoma \\
\midrule
GI & Colon adenocarcinoma, Stomach adenocarcinoma \\
\midrule
GU & Chromophobe renal cell carcinoma, Clear cell renal cell carcinoma, Mixed germ cell tumor, Papillary renal cell carcinoma, Prostatic adenocarcinoma, Renal oncocytoma, Seminoma, Testicular lymphoma, Urothelial carcinoma \\
\midrule
GYN & High-grade serous ovarian carcinoma, Uterine endometrioid carcinoma \\
\midrule
H\&N & Adenoid cystic carcinoma, Oropharyngeal squamous cell carcinoma, Papillary thyroid carcinoma \\
\midrule
Lung & Lung adenocarcinoma, Lung squamous cell carcinoma, Typical carcinoid tumor \\
\midrule
PHB & Cholangiocarcinoma, Hepatocellular carcinoma, Pancreatic adenocarcinoma \\
\midrule
Skin & Cutaneous melanoma \\
\bottomrule
\end{tabular}
\caption*{\textbf{Extended Data Table 5: Included diagnoses for PathQABench by organ}. GI: Gastrointestinal, GU: Genitourinary, GYN: Gynecology, 
H\&N: Head and Neck, PHB: Pancreaticohepatobiliary.}
\label{tab:pathqabench_private_mc_diag}
\end{table}

\begin{table}
\centering
\begin{tabular}{lp{12cm}}
\toprule
Organ & Multiple choice options \\
\midrule
Brain & Pilocytic astrocytoma, Ependymoma, Glioblastoma, Pleomorphic xanthoastrocytoma, Subependymal giant cell astrocytoma, Oligodendroglioma, Subependymoma, Ganglioglioma, Medulloblastoma, Neurocytoma \\
\midrule
GI & Colon adenocarcinoma, Stomach adenocarcinoma, Esophageal squamous cell carcinoma, Peutz-jeghers polyp, Sessile serrated adenoma, Hyerplastic polyp, Malt lymphoma, Ulcerative colitis, Neuroendocrine tumor, Gastrointestinal stromal tumor \\
\midrule
GU & Chromophobe renal cell carcinoma, Clear cell renal cell carcinoma, Papillary renal cell carcinoma, Renal oncocytoma, Prostatic stromal sarcoma, Urothelial carcinoma, Prostatic adenocarcinoma, Testicular lymphoma, Seminoma, Mixed germ cell tumor \\
\midrule
GYN and Breast & High-grade serous ovarian carcinoma, Low-grade serous ovarian carcinoma, Mucinous cystadenoma, Uterine endometrioid carcinoma, Invasive ductal carcinoma, Invasive lobular carcinoma, Phyllodes tumor, Cervical squamous cell carcinoma, Dcis, Paget disease \\
\midrule
Head \& Neck & Oropharyngeal squamous cell carcinoma, Acinic cell carcinoma, Adenoid cystic carcinoma, Mucoepidermoid carcinoma, Papillary thyroid carcinoma, Medullary thyroid carcinoma, Pleomorphic adenoma, Mucosal melanoma, Squamous papilloma, Nut carcinoma \\
\midrule
Lung & Lung adenocarcinoma, Lung squamous cell carcinoma, Typical carcinoid tumor, Atypical carcinoid tumor, Hamartoma of lung, Meningothelial-like nodule, Pneumocytoma, Small cell carcinoma, Large cell carcinoma, Large cell neuroendocrine carcinoma \\
\midrule
PHB & Hepatocellular carcinoma, Pancreatic adenocarcinoma, Cholangiocarcinoma, Pancreatic neuroendocrine tumor, Gallbladder adenocarcinoma, Cirrhosis, Hepatoblastoma, Intraductal papillary mucinous neoplasm, Mucinous cystic neoplasm, Fibrolamellar carcinoma \\
\midrule
Skin & Cutaneous melanoma, Basal cell carcinoma, Squamous cell carcinoma, Merkel cell carcinoma, Mycosis fungoides, Verruca vulgaris, Dermatofibroma, Blue nevus, Dysplastic nevus, Hidradenoma \\
\bottomrule
\end{tabular}
\caption*{\textbf{Extended Data Table 6: Options for multiple choice diagnostic questions in PathQABench by organ}. For each organ system, the pathologist selected a set of 10 possible answers that encompasses the correct answers for all questions within that organ system as well as other relatively common diagnoses within that organ system. Finally, when each multiple choice question is constructed using the above options, the order in each the options appear is randomized to ensure the correct answer is equally likely to be slotted into any location among the possible choices. GI: Gastrointestinal, GU: Genitourinary, GYN: Gynecology, 
H\&N: Head and Neck, PHB: Pancreaticohepatobiliary.}
\end{table}

\begin{table}
\centering
\begin{tabular}{lrr}
\toprule
Model & Combined & Combined w/ Context\\
\midrule
PathChat & \textbf{0.708 (0.583, 0.833)} & \textbf{0.812 (0.708, 0.917)} \\
LLaVA-Med & 0.188 (0.083, 0.292) & 0.271 (0.167, 0.396) \\
LLaVA 1.5 & 0.208 (0.104, 0.333) & 0.271 (0.166, 0.417) \\
\bottomrule
\end{tabular}
\caption*{\textbf{Extended Data Table 7: Performance on PathQABench multiple-choice diagnostic questions.} Accuracy is reported on the full set of PathQABench multiple choice questions ($n = 48$) in both the image only evaluation setting and the image + clinical context evaluation setting (denoted as ``w/ Context"). 95\% confidence intervals from bootstrapping are included in parentheses. For more details see \textbf{MLLM evaluation} in \textbf{Methods}.}
\label{tab:pathqabench_mc_combined}
\end{table}

\begin{table}
\centering
\begin{tabular}{lrr}
\toprule
Model & PathQABench-Public & PathQABench-Public w/ Context\\
\midrule
PathChat & \textbf{0.826 (0.652, 0.957)} & \textbf{0.870 (0.696, 1.000)} \\
GPT4V & $\ast$ 0.217 (0.043, 0.391) & 0.696 (0.478, 0.870) \\
LLaVA-Med & 0.130 (0.000, 0.261) & 0.174 (0.043, 0.348) \\
LLaVA 1.5 & 0.174 (0.043, 0.348) & 0.304 (0.130, 0.479) \\
\bottomrule
\end{tabular}
\caption*{\textbf{Extended Data Table 8: Performance on PathQABench-Public multiple-choice diagnostic questions.} Accuracy is reported on the PathQABench-Public multiple choice questions ($n = 23$) in both the image only evaluation setting and the image + clinical context evaluation setting (denoted as ``w/ Context"). 95\% confidence intervals from bootstrapping are included in parentheses. For more details see \textbf{MLLM evaluation} in \textbf{Methods}. $\ast$Note that due to guardrails implemented by GPT4V, only 12 / 23 questions submitted to the API yielded successful answers for PathQABench-Public in the image only setting (a maximum number of 3 attempts were made for each question). An unsuccessful query was treated as incorrect since the response did not address the question. We also report performance just on the subset of questions that yielded successful queries for GPT4V in \textbf{Extended Data Table 10}. In the image + clinical context setting, all questions were successfully answered by GPT4V. For more details see \textbf{Evaluation of GPT4V} in \textbf{Methods}.}
\label{tab:pathqabench_mc_public}
\end{table}

\begin{table}
\centering
\begin{tabular}{lrr}
\toprule
Model & PathQABench-Private & PathQABench-Private w/ Context\\
\midrule
PathChat & \textbf{0.600 (0.400, 0.800)} & \textbf{0.760 (0.600, 0.920)} \\
LLaVA-Med & 0.240 (0.080, 0.400) & 0.360 (0.160, 0.560) \\
LLaVA 1.5 & 0.240 (0.080, 0.400) & 0.240 (0.080, 0.440) \\
\bottomrule
\end{tabular}
\caption*{\textbf{Extended Data Table 9: Performance on PathQABench-Private multiple-choice diagnostic questions.} Accuracy is reported on the PathQABench-Private multiple choice questions ($n = 25$) in both the image only evaluation setting and the image + clinical context evaluation setting (denoted as ``w/ Context"). 95\% confidence intervals from bootstrapping are included in parentheses. For more details see \textbf{MLLM evaluation} in \textbf{Methods}.}
\label{tab:pathqabench_mc_private}
\end{table}

\begin{table}
\centering
\begin{tabular}{lrr}
\toprule
Model & PathQABench-Public & PathQABench-Public w/ Context\\
\midrule
PathChat & \textbf{0.667 (0.417, 0.917)} & \textbf{0.870 (0.696, 1.000)} \\
GPT4V & 0.417 (0.167, 0.750) & 0.696 (0.478, 0.870) \\
LLaVA-Med & 0.083 (0.000, 0.250) & 0.174 (0.043, 0.348) \\
LLaVA 1.5 & 0.167 (0.000, 0.417) & 0.304 (0.130, 0.479) \\
\bottomrule
\end{tabular}
\caption*{\textbf{Extended Data Table 10: Performance on PathQABench-Public multiple-choice diagnostic questions, restricted to sucessful GPT4V queries.} Accuracy is reported on the subset of PathQABench-Public multiple choice questions for which GPTV successfully answered the question (\textit{i.e.}, did not refuse to give an answer within the maximum number of 3 attempted API calls), for both the image only evaluation setting ($n = 12$) and the image + clinical context evaluation setting (denoted as ``w/ Context", $n = 23$). 95\% confidence intervals from bootstrapping are included in parentheses. For more details see \textbf{MLLM evaluation} in \textbf{Methods}.}
\label{tab:pathqabench_mc_public}
\end{table}

\begin{table}
\centering
\begin{tabular}{lr}
\toprule
Model & Accuracy \\
\midrule
PathChat & \textbf{0.861 (0.791, 0.922)} \\
GPT4V$\ast$ & 0.591 (0.504, 0.670) \\
LLaVA 1.5 & 0.426 (0.330, 0.513) \\
LLaVA-Med & 0.504 (0.417, 0.600) \\
\bottomrule
\end{tabular}
\caption*{\textbf{Extended Data Table 11: Proportion of open-ended questions in PathQABench correctly answered by each model}. 95\% confidence intervals from bootstrapping are included in parentheses. See \textbf{Extended Data Table 15 and 17} for accuracy stratified by category. $\ast$Note that due to guardrails implemented by GPT4V, only 97 / 115 questions submitted to the API yielded successful answers for PathQABench-Public (a maximum number of 3 attempts were made for each question). An unsuccessful query was treated as incorrect since the response did not address the question. We also report performance just on the subset of questions that yielded successful queries for GPT4V in \textbf{Extended Data Table 20}. For more details see \textbf{Evaluation of GPT4V} in \textbf{Methods}.}
\end{table}

\begin{table}
\centering
\begin{tabular}{lrrr}
\toprule
PathChat \textit{vs.} model & Lose & Tie & Win\\
\midrule
GPT4V$\ast$ & 0.296 (0.209, 0.374) & 0.130 (0.078, 0.191) & 0.574 (0.478, 0.661) \\
LLaVA-Med & 0.148 (0.087, 0.217) & 0.148 (0.087, 0.217) & 0.704 (0.617, 0.791) \\
LLaVA 1.5 & 0.113 (0.061, 0.174) & 0.191 (0.122, 0.270) & 0.696 (0.609, 0.783) \\
\bottomrule
\end{tabular}
\caption*{\textbf{Extended Data Table 12: Head-to-head performance of PathChat against other MLLMs on PathQABench open-ended questions.} For each competing model (GPT4V, LLaVA-Med, LLaVA 1.5), we compute the lose/tie/win rate ($n = 115$) of PathChat against said model. Lose: said model is ranked higher than PathChat; Tie: PathChat is tied with the model in ranking; Win: PathChat is ranked higher than the model. 95\% confidence intervals from bootstrapping are included in parentheses. For more details see \textbf{PathChat model evaluation} in \textbf{Methods}. $\ast$Note that due to guardrails implemented by GPT4V, only 97 / 115 queries to the API yielded successful answers for PathQABench-Public (a maximum number of 3 attempts were made for each question). An unsuccessful query was treated as incorrect (and therefore ranked as last or tied for last) since the response did not address the question. We also report performance just on the subset of questions that yielded successful queries for GPT4V in \textbf{Extended Data Table 23}. For more details see \textbf{Evaluation of GPT4V} in \textbf{Methods}.}
\label{tab:pathqabench_oe}
\end{table}

\begin{table}
\centering
\begin{tabular}{lp{12cm}}
\toprule
Broad Category & Description \\
\midrule
Microscopy &  Questions test the ability of models to generate accurate and detailed morphological descriptions of histology microscopy images and assess clinically relevant features such as tumor differentiation and grade \newline
Sub-categories: Microscopic description, Differentiation, Grading\\
\midrule
Diagnosis & Questions test the ability of models to directly suggest a reasonable diagnosis based on the histology image available and relevant clinical context  \newline
Sub-categories: Diagnosis\\
\midrule
Clinical & Questions test the ability of models to retrieve clinically relevant background knowledge about the disease in question, including risk factors, prognosis and treatment. \newline
Sub-categories: Risks, Prognosis, Treatment\\
\midrule
Ancillary testing & Questions test the ability of models to suggest additional testing such as IHCs and molecular to confirm a specific diagnosis \newline
Sub-categories: IHC, Molecular, Further testing\\
\bottomrule
\end{tabular}
\caption*{\textbf{Extended Data Table 13: Categorization of open-ended questions in PathQABench}. Number of questions in each category is summarized in \textbf{Extended Data Table 14}. Some questions may fit the description of more than one category or sub-category. Examples of each category can be found in \textbf{Extended Data Figure 5}.}
\end{table}

\begin{table}
\centering
\begin{tabular}{lrlr}
\toprule
Broad category & Count & Narrow category & Count\\
\midrule
\multirow{3}{*}{Microscopy} & \multirow{3}{*}{47} & Microscopic Description & 27\\
 & & Differentiation & 20\\
 & & Grading & 20\\
\midrule
Diagnosis & 23 & Diagnosis & 23\\
\midrule
\multirow{3}{*}{Clinical} & \multirow{3}{*}{26} & Risk Factors & 4\\
 & & Prognosis & 20\\
 & & Treatment & 22\\
\midrule
\multirow{3}{*}{Ancillary Testing} & \multirow{3}{*}{40} & IHC & 17\\
 & & Molecular & 21\\
 & & Other Testing & 4\\
\bottomrule
\end{tabular}
\caption*{\textbf{Extended Data Table 14: Broad and sub-categories for PathQABench open-ended questions.} In total of 115 questions were curated and reviewed by a board-certified anatomic pathologist from 23 cases in PathQABench-Public. Each question may fall under more than one category. }
\label{tab:pathqabench_oe}
\end{table}

\begin{table}
\centering
\begin{tabular}{lrrrr}
\toprule
Category & PathChat & GPT4V$\ast$ & LLaVA-Med & LLaVA 1.5 \\
\midrule
Microscopy & \textbf{0.830 (0.702, 0.936)} & 0.298 (0.170, 0.426) & 0.426 (0.298, 0.574) & 0.404 (0.277, 0.532) \\
Diagnosis & \textbf{0.739 (0.565, 0.913)} & 0.391 (0.217, 0.609) & 0.435 (0.260, 0.652) & 0.130 (0.000, 0.304) \\
Clinical & 0.923 (0.808, 1.000) & \textbf{1.000 (1.000, 1.000)} & 0.692 (0.500, 0.846) & 0.769 (0.614, 0.923) \\
Ancillary Testing & 0.925 (0.850, 1.000) & \textbf{0.975 (0.925, 1.000)} & 0.575 (0.425, 0.725) & 0.525 (0.350, 0.675) \\
\bottomrule
\end{tabular}
\caption*{\textbf{Extended Data Table 15: Proportion of open-ended questions in PathQABench correctly answered by each model, stratified by broad categories}. See \textbf{Extended Data Table 17} for accuracy stratified by category. $\ast$Note that due to guardrails implemented by GPT4V, only 97 / 115 questions submitted to the API yielded successful answers for PathQABench-Public (a maximum number of 3 attempts were made for each question). An unsuccessful query was treated as incorrect since the response did not address the question. We also report performance just on the subset of questions that yielded successful queries for GPT4V in \textbf{Extended Data Table 21}. For more details see \textbf{Evaluation of GPT4V} in \textbf{Methods}.}
\end{table}

\begin{table}
\centering
\begin{tabular}{llrrr}
\toprule
Category & PathChat \textit{vs.} model & Lose & Tie & Win\\
\midrule
\multirow{3}{*}{Microscopy} & GPT4V$\ast$ & 0.128 (0.043, 0.234) & 0.149 (0.064, 0.255) & 0.723 (0.596, 0.851) \\
  & LLaVA-Med & 0.128 (0.043, 0.234) & 0.213 (0.106, 0.319) & 0.660 (0.511, 0.787) \\
  & LLaVA 1.5 & 0.064 (0.021, 0.149) & 0.213 (0.106, 0.340) & 0.723 (0.596, 0.851) \\
\midrule
\multirow{3}{*}{Diagnosis} & GPT4V$\ast$ & 0.130 (0.043, 0.304) & 0.174 (0.043, 0.348) & 0.696 (0.478, 0.870) \\
  & LLaVA-Med & 0.130 (0.043, 0.261) & 0.174 (0.043, 0.316) & 0.696 (0.522, 0.870) \\
  & LLaVA 1.5 & 0.043 (0.043, 0.174) & 0.217 (0.086, 0.391) & 0.739 (0.565, 0.913) \\
\midrule
\multirow{3}{*}{Clinical} & GPT4V$\ast$ & 0.654 (0.500, 0.846) & 0.154 (0.038, 0.308) & 0.192 (0.077, 0.346) \\
  & LLaVA-Med & 0.231 (0.077, 0.423) & 0.154 (0.038, 0.308) & 0.615 (0.423, 0.808) \\
  & LLaVA 1.5 & 0.346 (0.154, 0.538) & 0.269 (0.115, 0.423) & 0.385 (0.192, 0.577) \\
\midrule
\multirow{3}{*}{Ancillary Testing} & GPT4V$\ast$ & 0.600 (0.425, 0.750) & 0.050 (0.025, 0.125) & 0.350 (0.200, 0.525) \\
  & LLaVA-Med & 0.175 (0.075, 0.300) & 0.025 (0.025, 0.098) & 0.800 (0.675, 0.900) \\
  & LLaVA 1.5 & 0.175 (0.075, 0.300) & 0.150 (0.050, 0.275) & 0.675 (0.525, 0.825) \\
\bottomrule
\end{tabular}
\caption*{\textbf{Extended Data Table 16: Head-to-head performance of PathChat against other MLLMs on PathQABench open-ended questions, stratified by broad categories.} For each competing model (LLaVA 1.5, LLaVA-Med, GPT4V), we compute the lose/tie/win rate of PathChat against said model. Within each category, Lose: said model is ranked higher than PathChat; Tie: PathChat is tied with the model in ranking; Win: PathChat is ranked higher than the model. 95\% confidence intervals from bootstrapping are included in parentheses. For more details see \textbf{PathChat model evaluation} in \textbf{Methods}. $\ast$Note that due to guardrails implemented by GPT4V, only 97 / 115 questions submitted to the API yielded successful answers for PathQABench-Public (a maximum number of 3 attempts were made for each question). An unsuccessful query was treated as incorrect (and therefore ranked as last or tied for last) since the response did not address the question. We also report performance just on the subset of questions that yielded successful queries for GPT4V in \textbf{Extended Data Table 24}. For more details see \textbf{Evaluation of GPT4V} in \textbf{Methods}.}
\label{tab:pathqabench_oe_broad}
\end{table}

\begin{table}
\centering
\begin{tabular}{lrrrr}
\toprule
Category & PathChat & GPT4V$\ast$ & LLaVA-Med & LLaVA 1.5 \\
\midrule
Microscopic & \textbf{0.852 (0.704, 0.963)} & 0.333 (0.148, 0.519) & 0.370 (0.185, 0.556) & 0.333 (0.148, 0.519) \\
Differentiation & \textbf{0.800 (0.600, 0.950)} & 0.250 (0.100, 0.450) & 0.450 (0.200, 0.651) & 0.450 (0.250, 0.650) \\
Grading & \textbf{0.750 (0.550, 0.950)} & 0.150 (0.000, 0.300) & 0.450 (0.250, 0.650) & 0.450 (0.250, 0.650) \\
Diagnosis & \textbf{0.739 (0.565, 0.913)} & 0.391 (0.217, 0.609) & 0.435 (0.260, 0.652) & 0.130 (0.000, 0.304) \\
Risk factors & \textbf{1.000 (1.000, 1.000)} & \textbf{1.000 (1.000, 1.000)} & 0.750 (0.250, 1.000) & 0.750 (0.250, 1.000) \\
Prognosis & 0.950 (0.850, 1.000) & \textbf{1.000 (1.000, 1.000)} & 0.750 (0.550, 0.950) & 0.850 (0.700, 1.000) \\
Treatment & 0.955 (0.864, 1.000) & \textbf{1.000 (1.000, 1.000)} & 0.727 (0.545, 0.909) & 0.773 (0.591, 0.955) \\
IHC & 0.882 (0.706, 1.000) & \textbf{0.941 (0.824, 1.000)} & 0.529 (0.294, 0.765) & 0.235 (0.059, 0.412) \\
Molecular & 0.952 (0.857, 1.000) & \textbf{1.000 (1.000, 1.000)} & 0.524 (0.332, 0.714) & 0.619 (0.429, 0.810) \\
Other Testing & \textbf{1.000 (1.000, 1.000)} & \textbf{1.000 (1.000, 1.000)} & 0.750 (0.250, 1.000) & \textbf{1.000 (1.000, 1.000)} \\
\bottomrule
\end{tabular}
\caption*{\textbf{Extended Data Table 17: Proportion of open-ended questions in PathQABench correctly answered by each model, stratified by sub-categories}. 95\% confidence intervals from bootstrapping are included in parentheses. $\ast$Note that due to guardrails implemented by GPT4V, only 97 / 115 questions submitted to the API yielded successful answers for PathQABench-Public (a maximum number of 3 attempts were made for each question). An unsuccessful query was treated as incorrect since the response did not address the question. We also report performance just on the subset of questions that yielded successful queries for GPT4V in \textbf{Extended Data Table 22}. For more details see \textbf{Evaluation of GPT4V} in \textbf{Methods}.}
\end{table}

\begin{table}
\centering
\begin{tabular}{llrrr}
\toprule
Category & PathChat \textit{vs.} model & Lose & Tie & Win\\
\midrule
\multirow{3}{*}{Microscopic} & GPT4V$\ast$ & 0.222 (0.074, 0.407) & 0.074 (0.037, 0.185) & 0.704 (0.519, 0.852) \\
  & LLaVA-Med & 0.148 (0.037, 0.296) & 0.148 (0.037, 0.296) & 0.704 (0.519, 0.852) \\
  & LLaVA 1.5 & 0.037 (0.037, 0.111) & 0.185 (0.037, 0.333) & 0.778 (0.593, 0.926) \\
\midrule
\multirow{3}{*}{Differentiation} & GPT4V$\ast$ & 0.000 (0.000, 0.000) & 0.250 (0.100, 0.450) & 0.750 (0.550, 0.900) \\
  & LLaVA-Med & 0.050 (0.000, 0.200) & 0.300 (0.100, 0.500) & 0.650 (0.450, 0.850) \\
  & LLaVA 1.5 & 0.100 (0.000, 0.250) & 0.200 (0.050, 0.400) & 0.700 (0.500, 0.900) \\
\midrule
\multirow{3}{*}{Grading} & GPT4V$\ast$ & 0.050 (0.050, 0.150) & 0.250 (0.100, 0.450) & 0.700 (0.500, 0.900) \\
  & LLaVA-Med & 0.150 (0.050, 0.300) & 0.300 (0.100, 0.500) & 0.550 (0.350, 0.800) \\
  & LLaVA 1.5 & 0.150 (0.050, 0.350) & 0.250 (0.100, 0.450) & 0.600 (0.400, 0.800) \\
\midrule
\multirow{3}{*}{Diagnosis} & GPT4V$\ast$ & 0.130 (0.043, 0.304) & 0.174 (0.043, 0.348) & 0.696 (0.478, 0.870) \\
  & LLaVA-Med & 0.130 (0.043, 0.261) & 0.174 (0.043, 0.316) & 0.696 (0.522, 0.870) \\
  & LLaVA 1.5 & 0.043 (0.043, 0.174) & 0.217 (0.086, 0.391) & 0.739 (0.565, 0.913) \\
\midrule
\multirow{3}{*}{Risk Factors} & GPT4V$\ast$ & 0.500 (0.000, 1.000) & 0.250 (0.000, 0.750) & 0.250 (0.000, 0.750) \\
  & LLaVA-Med & 0.250 (0.000, 0.750) & 0.250 (0.000, 0.750) & 0.500 (0.000, 1.000) \\
  & LLaVA 1.5 & 0.250 (0.000, 0.750) & 0.250 (0.000, 0.750) & 0.500 (0.000, 1.000) \\
\midrule
\multirow{3}{*}{Prognosis} & GPT4V$\ast$ & 0.650 (0.450, 0.850) & 0.100 (0.050, 0.250) & 0.250 (0.100, 0.450) \\
  & LLaVA-Med & 0.300 (0.100, 0.500) & 0.050 (0.000, 0.160) & 0.650 (0.449, 0.850) \\
  & LLaVA 1.5 & 0.450 (0.250, 0.650) & 0.150 (0.050, 0.300) & 0.400 (0.200, 0.600) \\
\midrule
\multirow{3}{*}{Treatment} & GPT4V$\ast$ & 0.682 (0.500, 0.864) & 0.136 (0.045, 0.273) & 0.182 (0.045, 0.364) \\
  & LLaVA-Med & 0.273 (0.091, 0.455) & 0.091 (0.045, 0.227) & 0.636 (0.455, 0.818) \\
  & LLaVA 1.5 & 0.364 (0.182, 0.591) & 0.273 (0.091, 0.455) & 0.364 (0.182, 0.545) \\
\midrule
\multirow{3}{*}{IHC} & GPT4V$\ast$ & 0.529 (0.235, 0.765) & 0.000 (0.000, 0.000) & 0.471 (0.235, 0.765) \\
  & LLaVA-Med & 0.176 (0.059, 0.353) & 0.000 (0.000, 0.000) & 0.824 (0.647, 1.000) \\
  & LLaVA 1.5 & 0.059 (0.059, 0.176) & 0.059 (0.000, 0.176) & 0.882 (0.706, 1.000) \\
\midrule
\multirow{3}{*}{Molecular} & GPT4V$\ast$ & 0.762 (0.571, 0.905) & 0.095 (0.048, 0.238) & 0.143 (0.048, 0.286) \\
  & LLaVA-Med & 0.190 (0.048, 0.352) & 0.048 (0.048, 0.143) & 0.762 (0.571, 0.952) \\
  & LLaVA 1.5 & 0.286 (0.095, 0.476) & 0.238 (0.048, 0.429) & 0.476 (0.286, 0.714) \\
\midrule
\multirow{3}{*}{Other Testing} & GPT4V$\ast$ & 0.000 (0.000, 0.000) & 0.000 (0.000, 0.000) & 1.000 (1.000, 1.000) \\
  & LLaVA-Med & 0.000 (0.000, 0.000) & 0.000 (0.000, 0.000) & 1.000 (1.000, 1.000) \\
  & LLaVA 1.5 & 0.000 (0.000, 0.000) & 0.000 (0.000, 0.000) & 1.000 (1.000, 1.000) \\
\bottomrule
\end{tabular}
\caption*{\textbf{Extended Data Table 18: Head-to-head performance of PathChat against other MLLMs on PathQABench open-ended questions, stratified by sub-categories.} For each competing model (LLaVA 1.5, LLaVA-Med, GPT4V), we compute the lose/tie/win rate of PathChat against said model. Within each category, Lose: said model is ranked higher than PathChat; Tie: PathChat is tied with the model in ranking; Win: PathChat is ranked higher than the model. 95\% confidence intervals from bootstrapping are included in parentheses. For more details see \textbf{PathChat model evaluation} in \textbf{Methods}. $\ast$Note that due to guardrails implemented by GPT4V, only 97 / 115 questions submitted to the API yielded successful answers for PathQABench-Public (a maximum number of 3 attempts were made for each question). An unsuccessful query was treated as incorrect (and therefore ranked as last or tied for last) since the response did not address the question. We also report performance just on the subset of questions that yielded successful queries for GPT4V in \textbf{Extended Data Table 25}. For more details see \textbf{Evaluation of GPT4V} in \textbf{Methods}.}
\label{tab:pathqabench_oe_narrow}
\end{table}

\begin{table}
\centering
\begin{tabular}{lrlr}
\toprule
Broad category & Count & Narrow category & Count\\
\midrule
\multirow{3}{*}{Microscopy} & \multirow{3}{*}{33 / 47} & Microscopic Description & 21 / 27\\
 & & Differentiation & 12 / 20\\
 & & Grading & 12 / 20\\
\midrule
Diagnosis & 19 / 23 & Diagnosis & 19 / 23\\
\midrule
\multirow{3}{*}{Clinical} & \multirow{3}{*}{26 / 26} & Risk Factors & 4 / 4\\
 & & Prognosis & 20 / 20\\
 & & Treatment & 22 / 22\\
\midrule
\multirow{3}{*}{Ancillary Testing} & \multirow{3}{*}{39 / 40} & IHC & 16 / 17\\
 & & Molecular & 21 / 21\\
 & & Other Testing & 4 / 4\\
\bottomrule
\end{tabular}
\caption*{\textbf{Extended Data Table 19: Broad and sub-categories for PathQABench open-ended questions succesfully answered by GPT4V} In total of 115 questions were curated and reviewed by a board-certified anatomic pathologist from 23 cases in PathQABench-Public. Each question may fall under more than one category. For each category, we indicate how many of the total questions were successfully answered by GPT4V within a maximum of 3 attempts (see \textbf{Evaluation of GPT4V} of \textbf{Methods}). }
\label{tab:pathqabench_oe}
\end{table}

\begin{table}
\centering
\begin{tabular}{lr}
\toprule
Model & Accuracy \\
\midrule
PathChat & \textbf{0.875 (0.802, 0.938)} \\
GPT4V & 0.708 (0.615, 0.792) \\
LLaVA-Med & 0.542 (0.448, 0.646) \\
LLaVA 1.5 & 0.458 (0.354, 0.562) \\
\bottomrule
\end{tabular}
\caption*{\textbf{Extended Data Table 20: Proportion of open-ended questions in PathQABench correctly answered by each model, restricted to successful GPT4V queries}. 95\% confidence intervals from bootstrapping are included in parentheses. See \textbf{Extended Data Table 21} for accuracy stratified by category. Accuracy is reported on the subset of PathQABench-Public open-ended questions ($n = 97$) for which GPT4V successfully answered the question (\textit{i.e.}, did not refuse to give an answer within the maximum number of 3 attempted API calls). For more details see \textbf{Evaluation of GPT4V} in \textbf{Methods}.}
\end{table}

\begin{table}
\centering
\begin{tabular}{lrrrr}
\toprule
Category & PathChat & GPT4V & LLaVA-Med & LLaVA 1.5 \\
\midrule
Microscopy & \textbf{0.812 (0.656, 0.938)} & 0.438 (0.281, 0.595) & 0.500 (0.312, 0.688) & 0.438 (0.281, 0.625) \\
Diagnosis & \textbf{0.789 (0.632, 0.947)} & 0.474 (0.263, 0.684) & 0.368 (0.158, 0.580) & 0.158 (0.000, 0.316) \\
Clinical & 0.923 (0.808, 1.000) & \textbf{1.000 (1.000, 1.000)} & 0.692 (0.500, 0.846) & 0.769 (0.614, 0.923) \\
Ancillary Testing & 0.923 (0.821, 1.000) & \textbf{1.000 (1.000, 1.000)} & 0.590 (0.436, 0.744) & 0.538 (0.385, 0.692) \\
\bottomrule
\end{tabular}
\caption*{\textbf{Extended Data Table 21: Proportion of open-ended questions in PathQABench correctly answered by each model, stratified by broad categories and restricted to successful GPT4V queries}. 95\% confidence intervals from bootstrapping are included in parentheses. See \textbf{Extended Data Table 22} for accuracy stratified by sub-category. Accuracy is reported on the subset of PathQABench-Public open-ended questions ($n = 97$) for which GPT4V successfully answered the question (\textit{i.e.}, did not refuse to give an answer within the maximum number of 3 attempted API calls). For more details see \textbf{Evaluation of GPT4V} in \textbf{Methods}.}
\end{table}

\begin{table}
\centering
\begin{tabular}{lrrrr}
\toprule
Category & PathChat & GPT4V & LLaVA-Med & LLaVA 1.5 \\
\midrule
Microscopic & \textbf{0.800 (0.600, 0.950)} & 0.450 (0.250, 0.650) & 0.350 (0.150, 0.550) & 0.400 (0.200, 0.650) \\
Differentiation & \textbf{0.833 (0.583, 1.000)} & 0.417 (0.167, 0.669) & 0.667 (0.417, 0.917) & 0.417 (0.167, 0.667) \\
Grading & \textbf{0.750 (0.500, 1.000)} & 0.250 (0.000, 0.500) & 0.667 (0.417, 0.917) & 0.500 (0.250, 0.750) \\
Diagnosis & \textbf{0.789 (0.632, 0.947)} & 0.474 (0.263, 0.684) & 0.368 (0.158, 0.580) & 0.158 (0.000, 0.316) \\
Risk factors & \textbf{1.000 (1.000, 1.000)} & \textbf{1.000 (1.000, 1.000)} & 0.750 (0.250, 1.000) & 0.750 (0.250, 1.000) \\
Prognosis & 0.950 (0.850, 1.000) & \textbf{1.000 (1.000, 1.000)} & 0.750 (0.550, 0.950) & 0.850 (0.700, 1.000) \\
Treatment & 0.955 (0.864, 1.000) & \textbf{1.000 (1.000, 1.000)} & 0.727 (0.545, 0.909) & 0.773 (0.591, 0.955) \\
IHC & 0.875 (0.688, 1.000) & \textbf{1.000 (1.000, 1.000)} & 0.562 (0.312, 0.812) & 0.250 (0.062, 0.500) \\
Molecular & 0.952 (0.857, 1.000) & \textbf{1.000 (1.000, 1.000)} & 0.524 (0.332, 0.714) & 0.619 (0.429, 0.810) \\
Other Testing & \textbf{1.000 (1.000, 1.000)} & \textbf{1.000 (1.000, 1.000)} & 0.750 (0.250, 1.000) & \textbf{1.000 (1.000, 1.000)} \\
\bottomrule
\end{tabular}
\caption*{\textbf{Extended Data Table 22: Proportion of open-ended questions in PathQABench correctly answered by each model, stratified by sub-categories and restricted to successful GPT4V queries}. 95\% confidence intervals from bootstrapping are included in parentheses. Accuracy is reported on the subset of PathQABench-Public open-ended questions ($n = 97$) for which GPT4V successfully answered the question (\textit{i.e.}, did not refuse to give an answer within the maximum number of 3 attempted API calls). For more details see \textbf{Evaluation of GPT4V} in \textbf{Methods}.}
\end{table}

\begin{table}
\centering
\begin{tabular}{lrrr}
\toprule
PathChat \textit{vs.} model & Lose & Tie & Win\\
\midrule
GPT4V & 0.351 (0.268, 0.454) & 0.113 (0.052, 0.175) & 0.536 (0.433, 0.629) \\
LLaVA-Med & 0.144 (0.082, 0.216) & 0.134 (0.072, 0.206) & 0.722 (0.629, 0.804) \\
LLaVA 1.5 & 0.113 (0.052, 0.175) & 0.206 (0.124, 0.299) & 0.680 (0.588, 0.773) \\
\bottomrule
\end{tabular}
\caption*{\textbf{Extended Data Table 23: Head-to-head performance of PathChat against other MLLMs on PathQABench open-ended questions, restricted to successful GPT4V queries} For each competing model (GPT4V, LLaVA-Med, LLaVA 1.5), we compute the lose/tie/win rate of PathChat against said model on the subset of questions ($n = 97$) for which GPT4V successfully answered the question (\textit{i.e.}, did not refuse to give an answer within the maximum number of 3 attempted API calls). Lose: said model is ranked higher than PathChat; Tie: PathChat is tied with the model in ranking; Win: PathChat is ranked higher than the model. 95\% confidence intervals from bootstrapping are included in parentheses. For more details see \textbf{MLLM model evaluation} in \textbf{Methods}. For more details see \textbf{Evaluation of GPT4V} in \textbf{Methods}.}
\label{tab:pathqabench_oe}
\end{table}

\begin{table}
\centering
\begin{tabular}{llrrr}
\toprule
Category & PathChat \textit{vs.} model & Lose & Tie & Win\\
\midrule
\multirow{3}{*}{Microscopy} & GPT4V$\ast$ & 0.182 (0.061, 0.333) & 0.152 (0.030, 0.273) & 0.667 (0.515, 0.818) \\
  & LLaVA-Med & 0.121 (0.030, 0.242) & 0.212 (0.091, 0.364) & 0.667 (0.515, 0.818) \\
  & LLaVA 1.5 & 0.030 (0.030, 0.091) & 0.303 (0.182, 0.455) & 0.667 (0.515, 0.818) \\
\midrule
\multirow{3}{*}{Diagnosis} & GPT4V$\ast$ & 0.158 (0.000, 0.316) & 0.105 (0.053, 0.263) & 0.737 (0.526, 0.895) \\
  & LLaVA-Med & 0.105 (0.000, 0.263) & 0.158 (0.053, 0.316) & 0.737 (0.578, 0.895) \\
  & LLaVA 1.5 & 0.053 (0.000, 0.158) & 0.158 (0.053, 0.316) & 0.789 (0.632, 0.947) \\
\midrule
\multirow{3}{*}{Clinical} & GPT4V$\ast$ & 0.654 (0.500, 0.846) & 0.154 (0.038, 0.308) & 0.192 (0.077, 0.346) \\
  & LLaVA-Med & 0.231 (0.077, 0.423) & 0.154 (0.038, 0.308) & 0.615 (0.423, 0.808) \\
  & LLaVA 1.5 & 0.346 (0.154, 0.538) & 0.269 (0.115, 0.423) & 0.385 (0.192, 0.577) \\
\midrule
\multirow{3}{*}{Ancillary Testing} & GPT4V$\ast$ & 0.615 (0.462, 0.744) & 0.051 (0.026, 0.128) & 0.333 (0.205, 0.487) \\
  & LLaVA-Med & 0.179 (0.077, 0.308) & 0.026 (0.026, 0.103) & 0.795 (0.667, 0.923) \\
  & LLaVA 1.5 & 0.179 (0.077, 0.308) & 0.154 (0.051, 0.282) & 0.667 (0.513, 0.821) \\
\bottomrule
\end{tabular}
\caption*{\textbf{Extended Data Table 24: Head-to-head performance of PathChat against other MLLMs on PathQABench open-ended questions, stratified by broad categories, restricted to successful GPT4V queries.} For each competing model (LLaVA 1.5, LLaVA-Med, GPT4V), we compute the lose/tie/win rate of PathChat against said model on the subset of questions ($n = 97$) for which GPT4V successfully answered the question (\textit{i.e.}, did not refuse to give an answer within the maximum number of 3 attempted API calls). Within each category, Lose: said model is ranked higher than PathChat; Tie: PathChat is tied with the model in ranking; Win: PathChat is ranked higher than the model. 95\% confidence intervals from bootstrapping are included in parentheses. For more details see \textbf{MLLM model evaluation} in \textbf{Methods}. For more details see \textbf{Evaluation of GPT4V} in \textbf{Methods}.}
\label{tab:pathqabench_oe_broad}
\end{table}

\begin{table}
\centering
\begin{tabular}{llrrr}
\toprule
Category & PathChat \textit{vs.} model & Lose & Tie & Win\\
\midrule
\multirow{3}{*}{Microscopic} & GPT4V$\ast$ & 0.286 (0.095, 0.476) & 0.095 (0.048, 0.238) & 0.619 (0.429, 0.810) \\
  & LLaVA-Med & 0.095 (0.048, 0.238) & 0.190 (0.048, 0.381) & 0.714 (0.524, 0.905) \\
  & LLaVA 1.5 & 0.048 (0.048, 0.143) & 0.238 (0.095, 0.429) & 0.714 (0.524, 0.905) \\
\midrule
\multirow{3}{*}{Differentiation} & GPT4V$\ast$ & 0.000 (0.000, 0.000) & 0.250 (0.083, 0.500) & 0.750 (0.500, 1.000) \\
  & LLaVA-Med & 0.083 (0.000, 0.250) & 0.250 (0.083, 0.500) & 0.667 (0.333, 0.917) \\
  & LLaVA 1.5 & 0.000 (0.000, 0.000) & 0.333 (0.083, 0.667) & 0.667 (0.333, 0.917) \\
\midrule
\multirow{3}{*}{Grading} & GPT4V$\ast$ & 0.083 (0.000, 0.250) & 0.250 (0.083, 0.500) & 0.667 (0.417, 0.917) \\
  & LLaVA-Med & 0.167 (0.000, 0.417) & 0.333 (0.083, 0.583) & 0.500 (0.250, 0.750) \\
  & LLaVA 1.5 & 0.083 (0.000, 0.250) & 0.417 (0.167, 0.667) & 0.500 (0.250, 0.750) \\
\midrule
\multirow{3}{*}{Diagnosis} & GPT4V$\ast$ & 0.158 (0.000, 0.316) & 0.105 (0.053, 0.263) & 0.737 (0.526, 0.895) \\
  & LLaVA-Med & 0.105 (0.000, 0.263) & 0.158 (0.053, 0.316) & 0.737 (0.578, 0.895) \\
  & LLaVA 1.5 & 0.053 (0.000, 0.158) & 0.158 (0.053, 0.316) & 0.789 (0.632, 0.947) \\
\midrule
\multirow{3}{*}{Risk Factors} & GPT4V$\ast$ & 0.500 (0.000, 1.000) & 0.250 (0.000, 0.750) & 0.250 (0.000, 0.750) \\
  & LLaVA-Med & 0.250 (0.000, 0.750) & 0.250 (0.000, 0.750) & 0.500 (0.000, 1.000) \\
  & LLaVA 1.5 & 0.250 (0.000, 0.750) & 0.250 (0.000, 0.750) & 0.500 (0.000, 1.000) \\
\midrule
\multirow{3}{*}{Prognosis} & GPT4V$\ast$ & 0.650 (0.450, 0.850) & 0.100 (0.050, 0.250) & 0.250 (0.100, 0.450) \\
  & LLaVA-Med & 0.300 (0.100, 0.500) & 0.050 (0.000, 0.160) & 0.650 (0.449, 0.850) \\
  & LLaVA 1.5 & 0.450 (0.250, 0.650) & 0.150 (0.050, 0.300) & 0.400 (0.200, 0.600) \\
\midrule
\multirow{3}{*}{Treatment} & GPT4V$\ast$ & 0.682 (0.500, 0.864) & 0.136 (0.045, 0.273) & 0.182 (0.045, 0.364) \\
  & LLaVA-Med & 0.273 (0.091, 0.455) & 0.091 (0.045, 0.227) & 0.636 (0.455, 0.818) \\
  & LLaVA 1.5 & 0.364 (0.182, 0.591) & 0.273 (0.091, 0.455) & 0.364 (0.182, 0.545) \\
\midrule
\multirow{3}{*}{IHC} & GPT4V$\ast$ & 0.562 (0.312, 0.812) & 0.000 (0.000, 0.000) & 0.438 (0.188, 0.688) \\
  & LLaVA-Med & 0.188 (0.062, 0.375) & 0.000 (0.000, 0.000) & 0.812 (0.625, 1.000) \\
  & LLaVA 1.5 & 0.062 (0.062, 0.250) & 0.062 (0.000, 0.188) & 0.875 (0.688, 1.000) \\
\midrule
\multirow{3}{*}{Molecular} & GPT4V$\ast$ & 0.762 (0.571, 0.905) & 0.095 (0.048, 0.238) & 0.143 (0.048, 0.286) \\
  & LLaVA-Med & 0.190 (0.048, 0.352) & 0.048 (0.048, 0.143) & 0.762 (0.571, 0.952) \\
  & LLaVA 1.5 & 0.286 (0.095, 0.476) & 0.238 (0.048, 0.429) & 0.476 (0.286, 0.714) \\
\midrule
\multirow{3}{*}{Other Testing} & GPT4V$\ast$ & 0.000 (0.000, 0.000) & 0.000 (0.000, 0.000) & 1.000 (1.000, 1.000) \\
  & LLaVA-Med & 0.000 (0.000, 0.000) & 0.000 (0.000, 0.000) & 1.000 (1.000, 1.000) \\
  & LLaVA 1.5 & 0.000 (0.000, 0.000) & 0.000 (0.000, 0.000) & 1.000 (1.000, 1.000) \\
\bottomrule
\end{tabular}
\caption*{\textbf{Extended Data Table 25: Head-to-head performance of PathChat against other MLLMs on PathQABench open-ended questions, stratified by sub-categories, restricted to sucessful GPT4V queries.} For each competing model (LLaVA 1.5, LLaVA-Med, GPT4V), we compute the lose/tie/win rate of PathChat against said model on the subset of questions ($n = 97$) for which GPT4V successfully answered the question (\textit{i.e.}, did not refuse to give an answer within the maximum number of 3 attempted API calls). Within each category, Lose: said model is ranked higher than PathChat; Tie: PathChat is tied with the model in ranking; Win: PathChat is ranked higher than the model. 95\% confidence intervals from bootstrapping are included in parentheses. For more details see \textbf{MLLM evaluation} in \textbf{Methods}.}
\label{tab:pathqabench_oe_narrow}
\end{table}

% microscopic description: MD
% differentiation: DI
% grading: G
% risk factors: R
% prognosis: P
% treatment: T
% diagnosis: D
% IHC: I
% Molecular: M
% Further testing: F

% Microscopy: MD+DI+G
% Diagnosis: D
% Clinical: R+P+T
% Further testing: I+M+F

\begin{nolinenumbers}
\clearpage
\section*{References} 
\vspace{2mm}

\begin{spacing}{0.9}
\bibliographystyle{naturemag}
\bibliography{sample}
\end{spacing}
\end{nolinenumbers}

\end{document}